\documentclass[journal]{IEEEtran}
\usepackage{cite}
\usepackage{amsthm}

\usepackage{fancyhdr}

\ifCLASSINFOpdf
   \usepackage[pdftex]{graphicx}
\else
\fi
\usepackage{caption}
\usepackage{subfigure}
\usepackage{booktabs}

\usepackage{amsmath}
\usepackage{amssymb, amsfonts}
\usepackage{bbm, bm}

\usepackage{multicol}  
\usepackage{multirow} 
\usepackage{algorithm, algorithmic}

\usepackage{array}
\usepackage{url}

\begin{document}
%
\title{Graph-attention-based Casual Discovery with Trust Region-navigated Clipping Policy Optimization}
%
%
%

\author{Shixuan~Liu,~Yanghe~Feng,~Keyu~Wu,~Guangquan~Cheng,~Jincai~Huang,~and~Zhong~Liu
\thanks{This work is supported by grants from National Natural Science Foundation of China (No.71701205, No.62001495) and Natural Science Foundation of Hunan Province (No.2020JJ5675).}
\thanks{S.Liu, Y.Feng (Correspondence Author), K.Wu, G.Cheng, J.Huang and Z.Liu are with the College of Systems Engineering, National University of Defense Technology, Changsha, 410073, P.R. China (e-mail: liushixuan@nudt.edu.cn, fengyanghe@nudt.edu.cn, keyuwu@nudt.edu.cn, chengguangquan@nudt.edu.cn, huangjincai@nudt.edu.cn, phillipliu@263.net).}
\thanks{Color versions of one or more figures in this article are available at
https://doi.org/10.1109/TCYB.2021.3116762.}
\thanks{
Digital Object Identifier 10.1109/TCYB.2021.3116762}
\thanks{© 2021 IEEE. Personal use of this material is permitted. Permission from IEEE must be
obtained for all other uses, in any current or future media, including
reprinting/republishing this material for advertising or promotional purposes, creating new
collective works, for resale or redistribution to servers or lists, or reuse of any copyrighted
component of this work in other works.}

}

\markboth{IEEE TRANSACTIONS ON CYBERNETICS}%
{Liu \textit{et al.}: Graph-attention-based Casual Discovery with Trust
Region-navigated Clipping Policy Optimization}

\maketitle

\begin{abstract}
In many domains of empirical sciences, discovering the causal structure within variables remains an indispensable task. Recently, to tackle with unoriented edges or latent assumptions violation suffered by conventional methods, researchers formulated a reinforcement learning (RL) procedure for causal discovery, and equipped REINFORCE algorithm to search for the best-rewarded directed acyclic graph. The two keys to the overall performance of the procedure are the robustness of RL methods and the efficient encoding of variables. However, on the one hand, REINFORCE is prone to local convergence and unstable performance during training. Neither trust region policy optimization, being computationally-expensive, nor proximal policy optimization (PPO), suffering from aggregate constraint deviation, is decent alternative for combinatory optimization problems with considerable individual subactions. We propose a trust region-navigated clipping policy optimization method for causal discovery that guarantees both better search efficiency and steadiness in policy optimization, in comparison with REINFORCE, PPO and our prioritized sampling-guided REINFORCE implementation. On the other hand, to boost the efficient encoding of variables, we propose a refined graph attention encoder called SDGAT that can grasp more feature information without priori neighbourhood information. With these improvements, the proposed method outperforms former RL method in both synthetic and benchmark datasets in terms of output results and optimization robustness.
\end{abstract}


%
\IEEEpeerreviewmaketitle

\section{Introduction}
%
%
%
%
\IEEEPARstart{T}{he} causal structure of the data generation process remains pivotal for many research issues. Ascertaining the causal mechanisms behind natural phenomenon are important in many scientific domains, e.g. we can hope to develop new drugs or prevent epidemic outbreak given the knowledge of virus mechanism\cite{Daniela2010}. Recent research has also shown that the integration of causal discovery is beneficial to semi-supervised learning and transfer learning tasks, e.g. in function estimation cases with inferred causality\cite{Bernhard2019}. Although controlled randomized experiment is the common approach for causality discovery, yet in some scientific domains, it is almost impossible to simulate such experiment\cite{Rainer2007}. In this sense, recent researches in causal discovery methods focus on inferring causality from passively observable data\cite{Peter2000}.

Score-based methods formulate such structure discovery problem as adjusting the graph adjacency matrix to minimize predefined score function with acyclicity constraints. However, given the gigantic search space, whose complexity would rocket exponentially as the number of nodes grows, the optimization problem remains hideously tough to solve\cite{Chickering1995}. Fascinated by the search efficiency and ability to incorporate multiple indifferentiable score functions and constraints, a reinforcement learning (RL) procedure\cite{Zhu2020} was formulated recently to search for the best-scored directed acyclic graph (DAG) with Bayesian information criterion (BIC)\cite{Schwarz1978} as the score function using an encoder-decoder network architecture. The insight is, after graph generation using the observed data, a specific reinforcement learning agent optimize its policy w.r.t a stream of reward signals that comprised of the score function and acyclicity constraints calculated for the generated graph. The overall performance of such RL procedure relies heavily on the efficiency of RL methods and the compact encoding of variables that best capture the intrinsic relations.

Firstly, from a RL perspective, this paradigm has not reached its ultimate potential as the adopted REINFORCE\cite{Williams1992,Sutton1999} algorithm is susceptible to local convergence and poor search efficiency. Although trust region policy optimization (TRPO)\cite{Schulman2015} can guarantee a reliable and steady performance in resolving these issues, it is computationally expensive even with some approximation methods like conjugate gradient. On the other hand, even though proximal policy optimization (PPO)\cite{Schulman2017} entails some of the benefits of TRPO and is simpler to implement, yet we find that the deviation from trust region constraints, caused by trivially-bounded clipping, shall bring exponentially-large aggregate deviation after product calculation during training when the action is comprised of considerable subactions. This results in aberrant exploratory behaviours as the agent cannot optimize some sub-action or make them fall in local optima, particularly when they are not favored by the old policy.

Secondly, despite Graph Attention Network (GAT)\cite{Velickovic2017} outperforms other state-of-the-art graph neural networks in both transductive and inductive graph benchmark test, it still fails to provide a robust causal understanding in our task as the simple additive mechanism cannot capture the interrelations among variables without prior adjacency information.

Our contributions can be summarized as,
\begin{itemize}	
\item We propose Trust Region-navigated Clipping Policy Optimization (TRC) to help agent search over the directed graph space, without suffering from local convergence or aberrant search behaviour using REINFORCE or PPO. This insight is also applicable to other high dimensional combinatorial optimization problems. We also implement Prioritized Sampling-guided REINFORCE (PSR) that also show better convergence speed and output results than REINFORCE.

\item We leverage recent advancements in graph neural networks and scaled dot-product attention mechanism to form the scaled dot-product graph attention network (SDGAT) that have better causality extraction ability from the original data to boost accurate DAG generation.
\end{itemize}

We test our methods on both synthetic and benchmark datasets. TRC shows better performance in terms of causality discovery results and convergence speed than the REINFORCE and PSR, whilst SDGAT outperforms GAT and Transformer in terms of causality extraction ability and convergence stability respectively.

In what follows, Section \ref{sec2} discusses the related prior works. Section \ref{sec3} shall clarify the causality model and give a formulation of the RL process. The proposed graph attention network and RL methods are explained in detail in Section \ref{sec4} and Section \ref{sec5} respectively whilst their performance is examined in Section \ref{sec6}. We end our paper with conclusion in Section \ref{sec7}.

\section{Related Work}
\label{sec2}
\noindent Traditional causal discovery methods are comprised of score-based, constraint-based and hybrid methods. The prevailing score-based methods rely on pre-defined score functions to model the causal problem as searching over the DAGs space for the best-scored DAG. However, this problem remains NP-hard\cite{Chickering1995} to solve, therefore for practical problems with sizable node set, approximate search with extra structure assumption is often adopted.\cite{Chow1968} There are also hybrid methods that reduce the score-based search space given some assumed constraints\cite{Tsamardinos2006}, but these methods lack formalization on score functions and heuristics strategies.

Zheng et al.\cite{Zheng2018} formulate an equivalent acyclicity constraint with a continuous function of the adjacency matrix, which change the combinatorial nature of the problem to a continuous optimization problem. Despite this optimization problem only has stationary-point solutions rather than global optimum given the nonlinear nature, such local solutions are empirically highly comparable to the global solutions with expensive combinatorial searches. However, many effective score functions, e.g. generalized score function\cite{Huang2018} and independence-based score functions\cite{Peters2014}, cannot be incorporate with this approach as they either can be far too complicated or cannot be represented in closed forms.

Spurred by the applications of machine learning\cite{Ahmed2019AML} and deep learning\cite{9262059} in industrial systems\cite{preitl2007} and traffic management\cite{KatriniokA16}, recent advancement in neural networks also gives rise to the surface of neural-network-based causal discovery approaches. To learn the generative model of the joint distribution, Causal Generative Neural Network\cite{Goudet2017}, which is trained by minimizing the maximum mean discrepancy, is proposed. Kalainathan et al.\cite{Kalainathan2018} present a GAN-style method called Structural Agnostic Modelling, to recover full causal models from continuous observational data in an adversarial way. Still, Yu et al.\cite{Yu2019} propose DAG-GNN to generate DAG with an innovative graph neural network (GNN) architecture.

To extend neural networks to deal with arbitrarily-structured graphs, graph neural networks were introduced in Gori et al.\cite{Gori2005} and Scarselli et al.\cite{Scarselli2009}, to directly deal with a general class of graphs. Nevertheless, as an attempt to generalize convolutions to the graph domain, both spectral \cite{Tang2019} and non-spectral \cite{Hamilton2017} approaches are proposed. Nevertheless, as attention mechanisms have de facto become a norm in many sequence-to-sequence problems, Veličković et al.\cite{Velickovic2017} incorporate self-attention into the propagation step, achieving decent performance on both transductive and inductive tasks. Graph embedding

Reinforcement Learning (RL) methods also spur many applications in recent decades\cite{9043893}. Apart from its mostly renowned application in gaming, RL also shows decent applicability, robustness and generalization in robotics\cite{Polydoros2017, 8613842}, object localization\cite{Wang2019} and industrial processes\cite{8906005}. As an important insight for this paper, RL is also adopted to carry out neural architectural search, and achieved human-level results with robustness\cite{Zoph2016}.

\section{Preliminaries}
\label{sec3}
\noindent Causal graphs are all DAGs, and the objective of this paper is to search for the best-scored DAG $\mathcal G$ that best describes the data generation process:
\begin{equation}
\min_{\mathcal G\in DAGs} S(\mathcal G)
\end{equation}
Before we detail the graph generation specifications and learning methods, we specify the causal model adopted in this paper and formalize RL w.r.t the decision process as they are the foundations for further studies.
\subsection{Causal Model Definition}
Consider a finite observed random variables $\bold X =(\vec{X_i})_{i=1,...,n}^\mathrm{T}$, each scalar variable $x_i \in \vec{X_i}$ is associated with a node $i$ in graph $\mathcal G=(\bold V,\mathcal E)$ that consists of $n$ nodes $\bold V$ and edges $\mathcal E \subseteq \bold V^2$. A node $i$ can be regarded as a parent of $j$ if $(i,j)\in \mathcal E$. For any $v \in \bold V$, $(v,v) \not\in \mathcal E$. In this paper, we consider the data generation procedure as a DAG-based additive noise model, in which the observed value of $x_i$ is calculated by a function $\it f_i$ with variables on its parents node set $\bold{PA}_i$ in $\mathcal G$ as inputs, along with an independent additive noise $N_i$,

\begin{equation}
x_i=\it f_i(\bold{PA}_i)+N_i,\, i=1,...,n
\label{gen1}
\end{equation}

It is assumed that all noise variables $N_i$ have a strictly positive density and are jointly independent and thereby causal minimality reduces to that each function $\it f_i$ is not constant in any of its arguments\cite{Peters2014}. The above model can only be up to Markov equivalence class (DAGs set that encode the identical conditional independence statements) under usual Markov and faithfulness assumptions. To evaluate the applicability of the model, our  experiments on synthetic datasets are conducted with these further identifiable model settings:

\begin{itemize}
\item Linear Gaussian model assuming linear $\it f_i$ and Gaussian noises $N_i$\cite{Bollen1989,Spirtes2000}
\item Linear non-Gaussian acyclic model (LiNGAM) hypothesizing linear $\it f_i$ and non-Gaussian noises $N_i$\cite{Shimizu2006}
\item Non-linear model with quadratic $\it f_i$ using sampled coefficients and non-Gaussian noises $N_i$\cite{Peters2014}
\item Non-linear model with $\it f_i$ sampled from Gaussian-process (GP) and normally-distributed noises $N_i$\cite{Peters2014}
\end{itemize}
All the synthetic datasets are generated according to the above models with a fixed randomly-generated DAG and the dataset is sampled as $\bold X \in \mathbb R^{n\times M}$, where $M$ denotes the number of entries in the sampled dataset.

\subsection{Reinforcement Learning Formalization}

\begin{figure}[htbp]
 \centerline{\includegraphics [scale=0.57]{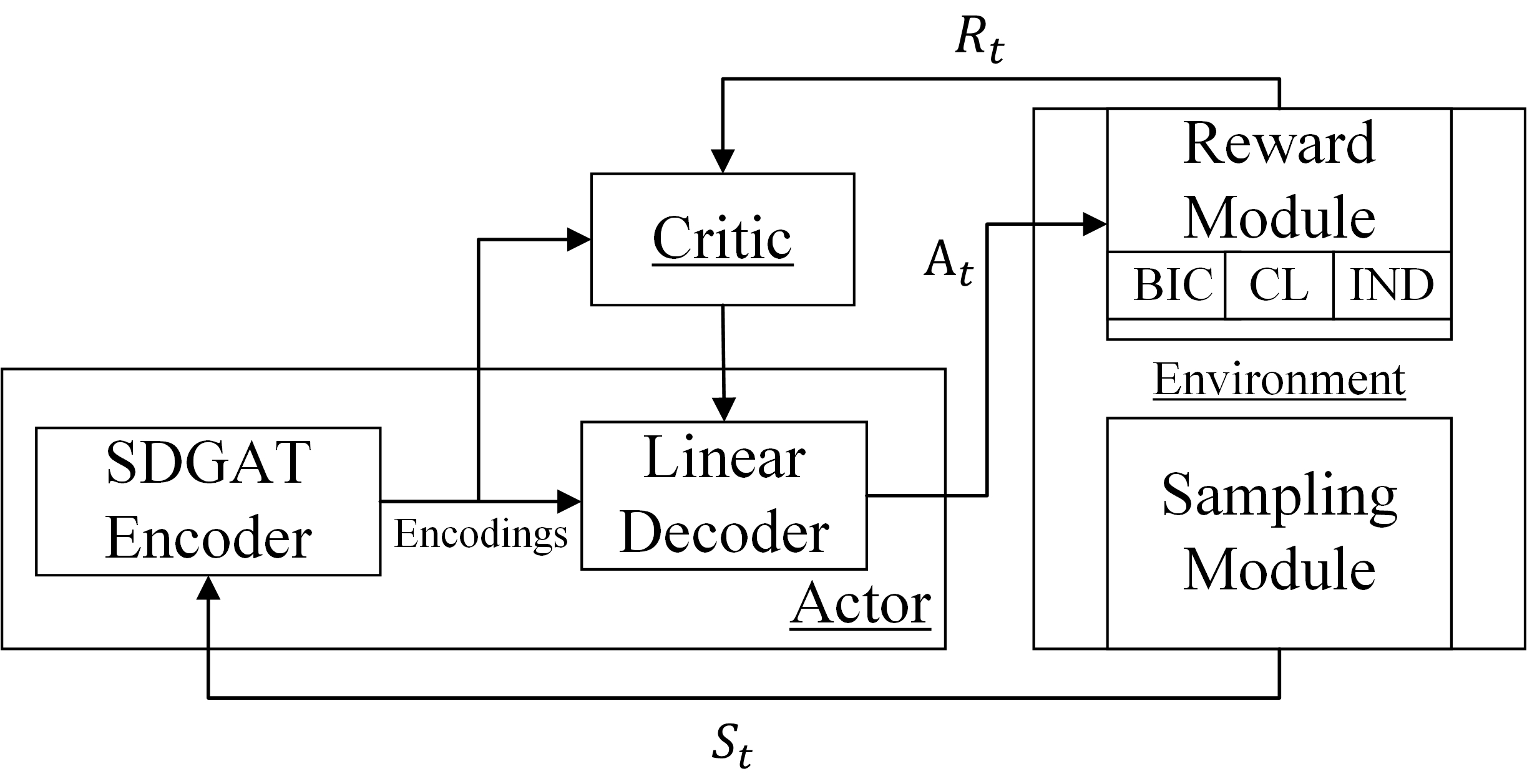}}
 \caption{RL Paradigm for Causal Discovery}
 \label{fig1}
\end{figure}
\begin{figure}[htbp]
  \centering
  \includegraphics [width=9cm,height=5.2cm]{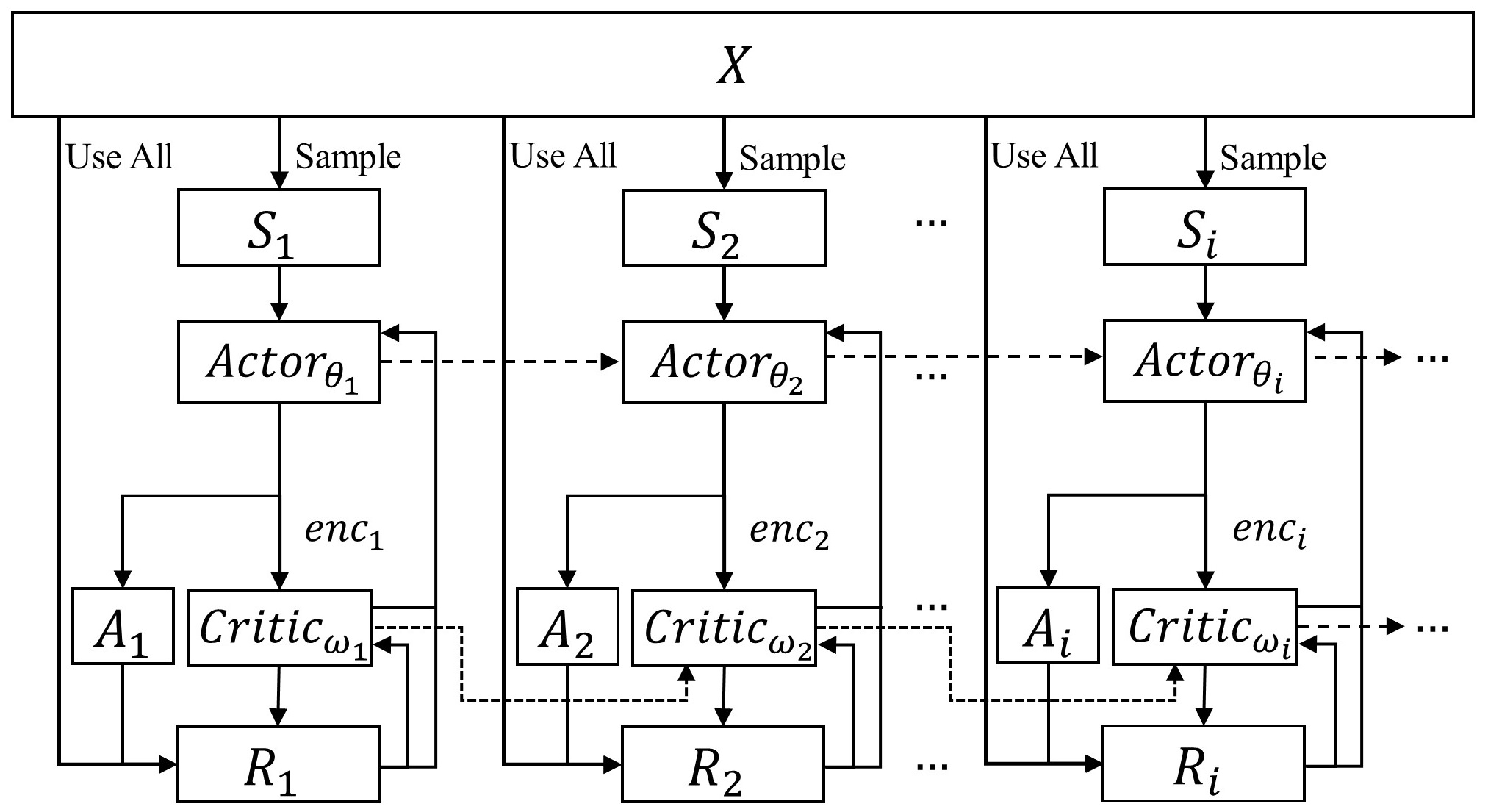}
  \caption{Intuitive Training Illustration with REINFORCE}
  \label{fig2}
\end{figure}

The RL model for causal discovery is shown in Fig. \ref{fig1}, which follows the actor-critic pattern. With this model, the objective of causal discovery is to discover the hidden causal DAG by continuously using the sampled variables $S_t$ from the observed data X. Based on an encoder-decoder module, the actor generates the adjacency matrix for graph $\mathcal G$ according to $S_t$. Given the actor output $A_t$, the reward module then calculates the reward $R_t$ for the critic to estimate the value function $V(S_t)$, with which $R_t$ forms the advantage signal that optimize the actor. To give a basic insight of the learning process, a training illustration using REINFORCE is shown in Fig. \ref{fig2}.

\subsubsection {Samples, Actions and Policy}
\label{fspa}
\begin{itemize}
\item 
$S\in \mathbb{R}^{n\times m}$ denotes the sample where $m$ is the feature depth for the sample. During training, $S_t$ is randomly sampled from the whole observed dataset $\bold X$ at time $t$. Since the dataset $\bold X$ does not change over time, the probability of $S_t$ being sampled remains the same.

\item 
$A\in\{{0,1}\}^{n\times n}$ denotes the action and can be understood as the generated binary adjacency matrix. At time $t$, $A_t=(a_t^{i,j})_{i,j=1...n}$, and its subaction $a_t^{i,j}=1$ implies that there is an edge from node $i$ to node $j$ at time $t$. Each subaction $a_t^{i,j}$ is generated according to a Bernouli distribution with the value of a given subpolicy. 

\item 
Being parameterized by $\theta$,  the policy $\pi_\theta(\cdot|S_t)=[\pi_{\theta}(a_t^{i,j}|S_t)]_{i,j=1...n}$ represents the adjacency probability matrix whose element-subpolicy $\pi_{\theta}(a_t^{i,j}|S_t) \in [0,1]$ implies the probability that $a_t^{i,j}=1$. To avoid self-loop, all the diagonal entries in the policy, during the last step of graph generation, are masked to zeros, i.e. $\pi_{\theta}(a_t^{i,i}|S_t)=0\,(i=1...n)$. However, even masking out the diagonal entries, the combination space for subactions, being $2^{n\times(n-1)}$, is still hideously high-dimensional, which is a source of local convergence and gradient overflow. As the subactions are sampled independently, the probability of $A_t$ being sampled in state $S_t$ is, 
\label{pass1}
\begin{equation}
P_{\theta}(A_t|S_t)=\prod_{i,j}^{n}\pi_{\theta}(a_t^{i,j}|S_t)
\end{equation}

\end{itemize}

\subsubsection {Rewards}
$R\in\mathbb{R}$ is the reward signal which incorporates both score function and acyclicity constraints. Traditional score-based methods adopt a parametric model for causality, which introduces a set of parameters $\mu$. 

Bayesian information criterion (BIC) is an asymptotic metric that measures the fitness of estimated models on a fixed dataset with identical numerical values for dependent variables and model with the lowest value of BIC is preferred. We leverage BIC as the score function for its consistency and decomposability. 
The BIC score for adjacency matrix $A_t$ given the whole dataset $X$ is,

\begin{equation}
BIC\left(A_t\right)=-2\,\ln\,p\left(X;\hat{\mu},A_t\right)+d_\mu \ln M
\end{equation}

where $\hat{\mu}$ denotes the maximum likelihood estimator parametrized by $\mu$ and $d_\mu$ is the dimensionality of $\mu$. If each causal relation is modelled using linear models, the BIC score can be given as,

\begin{equation}
{BIC}_1\left(A_t\right)=\sum_{i=1}^{n}{(M\ast \ln(\frac{\sum_{j=1}^{M}{(X_{i,j}-\widehat{X_{i,j}^{A_t})}}^2}{M}}))+n_e \ln M
\label{bic1}
\end{equation}

where $\widehat{X_{i,j}^{A_t}}$ denotes the estimate for $X_{i,j}$, the $i$-th entry in the k-th sample, using entries from parent node sets as indicated by $A_t$. $n_e$ is the number of edges, and the second term is therefore used to penalise edges redundancy. If we further assume equal additive noise variances among the node data, this can be rewritten as,

\begin{equation}
{BIC}_2\left(A_t\right)=Mn\ast \ln \sum\limits_{i=1}^{n} (\frac{\sum\nolimits_{j=1}^{M}{(X_{i,j}-\widehat{X_{i,j}^{A_t})}}^2}{Mn})+n_e \ln M
\label{bic2}
\end{equation}

Besides linear regression models, quadratic regression and GP regression are also used in this paper to for nonlinear data models accordingly to model causal relations given $X$. As for the cyclicity penalty term, given the insight of the acyclicity constraint proved in Zheng et al\cite{Zheng2018} that $tr\left(e^{A_t}\right)-n=0$ if $A_t$ is a directed graph, it is defined as,

\begin{equation}
CL(A_t)=tr\left(e^{A_t}\right)-n, \ e^{A_t}=\sum\nolimits_{k=0}^{\infty}{\frac{{A_t}^k}{k!}}
\end{equation}

However, for certain cyclic graph, $CL(A_t)$ is rather small and the minimum of all non-DAGs is hard to compute. Therefore, a large indicator penalty is added to induce particular DAGs with regard to acyclicity as,
\begin{equation}Ind\left(A_t\right)=\mathbbm{1}_{A_t\not\in DAGs}\end{equation}
As our objective is to minimize these three items, whilst RL typically deals with reward maximization, we formally define our training objective as,
\begin{equation}
\begin{split}
J(\theta|\bold S)&=\mathbbm{E}_{A \sim \pi_{\theta}(\cdot |\bold S)}\{R\}
\\
&=\mathbbm{E}_{A \sim \pi_{\theta}(\cdot |\bold S)}\{-[BIC\left(A\right)+\lambda_1CL\left(A\right)+\lambda_2Ind\left(A\right)]\}
\label{con0}
\end{split}
\end{equation}

\section {Neural Network Architecture for Graph Generation}
\label{sec4}
\noindent To discover the causality relations that best describes the generation process described in Equation (\ref{gen1}) from the observed data, we need to design a proper neural network architecture for graph generation. Encoder-decoder architecture has been the top priority for many sequence-to-sequence tasks like machine translation\cite{Cho2014} and text summarization\cite{Nallapati2016}, and it is also adopted in this paper.  As shown in Fig. \ref{fig1}, encoder and decoder together form the $actor_\theta$ module parameterized by $\theta$ that outputs the graph adjacency matrix $A_t$ given $S_t$. The encoder shall comprehend the intrinsic relation among variables and output an encoding $enc_i$ that best describes causality whilst the decoder shall use the encodings to interpret the interrelations among variables.
\subsection {Scaled Dot-Product Graph Attention Encoder}
\begin{figure}[htbp]
  \centering
  \includegraphics [scale=0.42]{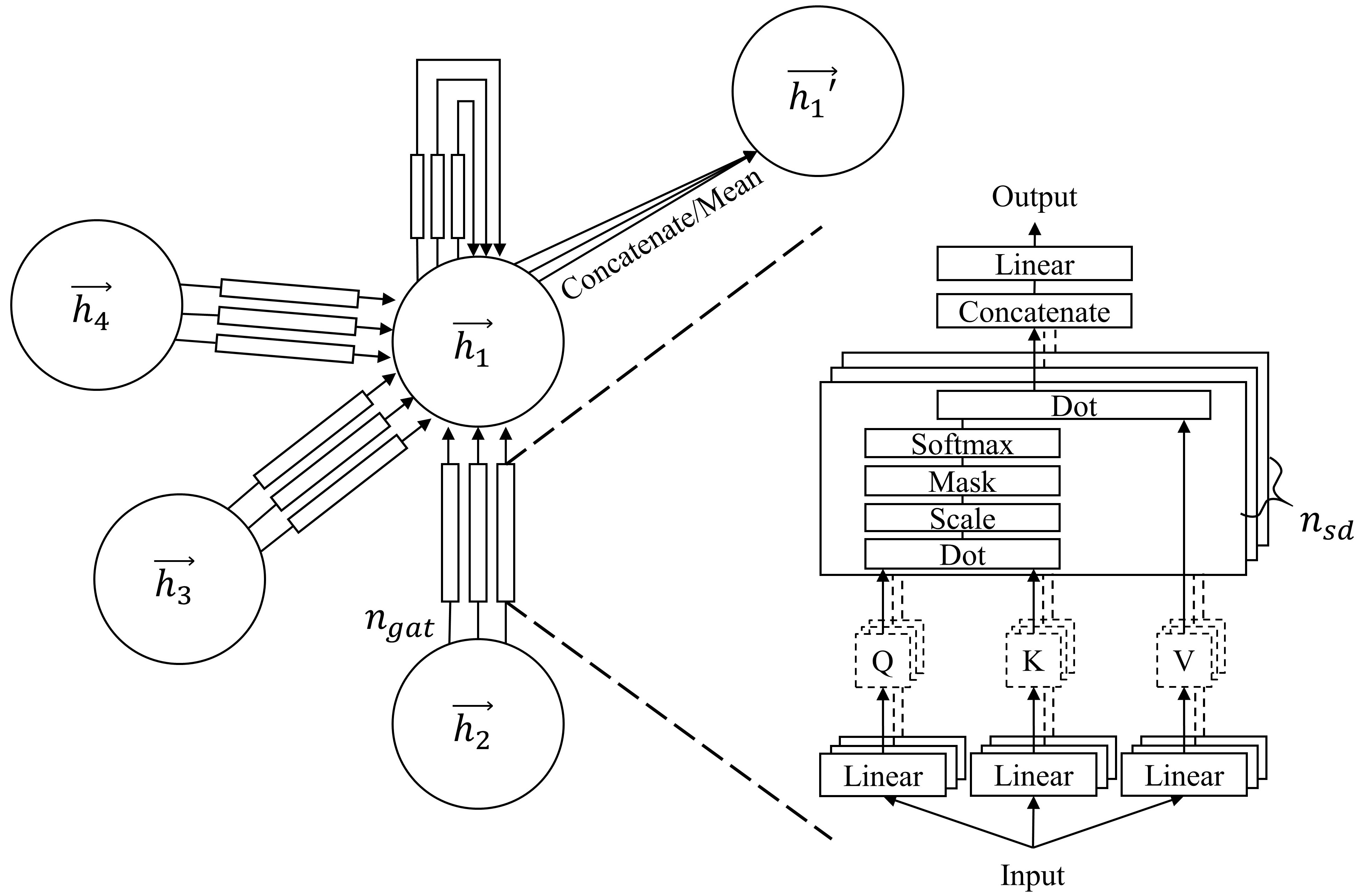}
  \caption{Scaled Dot-Product Graph Attention Network}
  \label{fig3}
\end{figure}

In order to provide sufficient interactions among variables, the design of GAT is often adopted by recent researches. However, in our experiment, its performance turns out to be miserable in terms of the generated graph and convergence speed, which will be analyzed in the experiment section. As there is no priori knowledge of the graph structure, the original model, ignoring all neighbourhood status, computes the mutual information among all node pairs. The two most commonly adopted attention functions are additive attention and multiplicative attention, and we note that the additive attention adopted in GAT cannot efficiently apprehend the interrelations among variables without adjacency information. 

We replace the original additive attention in GAT with scaled dot-product attention and they together form the SDGAT. Similar to GAT, SDGAT is also stacked by $n_s$ single attentional layers. For a set of $n$-node variables with $m$ features, $\bold h=\{\vec{h_1}, \vec{h_2}, ..., \vec{h_n}\}, \vec{h_i}\in \mathbbm{R}^m$, a single layer shall output a new set of $n$-node variables of cardinality $m'$.

As shown in Fig. \ref{fig3}, a two-hierarchical multi-head designs is employed in SDGAT. The first hierarchy resembles the counterpart in GAT whilst the second follows the pattern in Transformers. In the first hierarchy, $\bold h$ is duplicated as $n_{gat}$ copies, each represented by $\bold h(p) \in \mathbbm{R}^{n\times m}$ and indexed by $p$. Each $\bold h(p)$, is then plugged in parallel into the second hierarchy that has $n_{sd}$ heads, each of which follow the basic structure of scaled dot-product attention. In the $(p,q)$-th sublayer, the query $Q_{pq}$, key $K_{pq}$ and value $V_{pq}$ are obtained after plugging $\bold h(p)$ into three sets of independent linear projections,

\begin{equation}
Q_{pq}=\bold h(p)W_{pq}^Q, \ 
K_{pq}=\bold h(p)W_{pq}^K, \ 
V_{pq}=\bold h(p)W_{pq}^V
\end{equation}

where $W_{pq}^Q, W_{pq}^K \in \mathbbm{R}^{m \times d_k}, W_{pq}^V \in \mathbbm{R}^{m \times \frac{d_v}{n_{gat}\times n_{sd}}}$, and $d_k$ is the hidden dimension size for $Q_{pq}$ and $K_{pq}$. With corresponding queries and keys, all the attention matrices $\bm{\alpha}^{pq} \in \mathbbm{R}^{n\times n}$ for the $(p,q)$-th sublayer are computed simultaneously as,

\begin{equation}
\begin{split}
\bm{\alpha}^{pq}&=softmax(\frac{Q_{pq}{K_{pq}}^\mathrm{T}}{\sqrt{d_k}})
\\
&=softmax(\frac{\bold h(p)W_{pq}^Q{(W_{pq}^K)}^\mathrm{T}{\bold h(p)}^\mathrm{T}}{\sqrt{d_k}})
\end{split}
\end{equation}

where $\alpha_{ij}^{pq} \in \bm \alpha^{pq}$ indicates the importance level of node $j$'s features to node $i$ and $\frac{1}{\sqrt{d_k}}$ is the corresponding scaling factor. In the current setting, the model allows every node to get connected to other nodes as there is no initial structural information before graph generation. However, given existed graph structure, we could inject structure information into the attention mechanism by masking out certain $\alpha_{ij}^{pq}$ w.r.t the binary adjacency matrix. 

The $n_{sd}$ independent feature outputs are concatenated and are linearly projected before another concatenation for the $n_{gat}$ heads, resulting in the following output feature representation,
\begin{equation}
\vec{h_i'}=\mathop{\|} \limits_{p=1}^{n_{gat}}(\mathop{\|} \limits_{q=1}^{n_{sd}} \sum_{j\in {N_i}}\alpha_{ij}^{pq}\bold h(p)W_{pq}^V)W_p^O
\end{equation}
where $W_p^O \in \mathbbm{R}^{\frac{d_v}{n_{gat}} \times \frac{m'}{n_{gat}}}$. In our implementation, all the hidden dimensions and layer output dimensions remain the same, and after $n_s$ stacks, the encodings $enc \in \mathbbm{R}^{n \times m'}$ is obtained for the following graph generation process.

\subsection {Linear Graph Generation Decoder}
Given the encoding output $enc$, the linear graph generation decoder generate the graph adjacency matrix element-wise. By looping all $enc_i$-$enc_j$ pairs in $enc$, the subpolicy $\pi_{\theta}(a_t^{i,j}|S_t)$ discussed in \ref{pass1} is given as,

\begin{equation}
\pi_{\theta}(a_t^{i,j}|S_t)=sigmoid(q^{\mathrm{T}}tanh(W_1enc_i+W_2enc_j))
\end{equation}

$W_1,W_2\in \mathbbm{R}^{d_h\times m'}, q\in \mathbbm{R}^{d_h\times 1}$ are trainable variables where $d_h$ is the hidden dimension. All the diagonal entries are masked to 0. With each subaction $a_t^{i,j}$ sampled according to a Bernouli distribution with probability $\pi_{\theta}(a_t^{i,j}|S_t)$, the binary adjacency matrix $A_t$ is generated.

\section {Reinforcement Learning Algorithms for Causal Discovery}
As we leverage RL agent to search over the DAG space, its search efficiency shall be optimized to its best. In this section, we specify the traditional and proposed RL methods that boost its search ability.

\label{sec5}
\subsection {REINFORCE with Moving Average Baseline}
\begin{algorithm}[htbp]
\caption{REINFORCE with Moving Average Baseline}
\begin{algorithmic}[1] 
\REQUIRE Batch size $N$; Moving average update rate $\alpha_{m}$; Learning rates for actor and critic $\alpha_\theta, \alpha_\omega$; Entropy term weight $\lambda_e$;
\FOR{$t=1,2,...$}
\STATE Generate online experience batch $\{S_t, A_t, R_t\}_N$

\STATE $R_m\leftarrow (1-\alpha_{m})\overline{\{R_t\}_N}+\alpha_{m}R_m$

\STATE $\{\hat{A_t}\}_N\leftarrow \{R_t\}_N-R_m-V_\omega(\{S_t\}_N)$

\STATE $L_{\theta}\leftarrow \overline{\{{\hat{A_t}*[\frac{1}{n^2}\sum\limits_{i,j}\ln(\pi_{\theta}(A_t^{i,j}\mid S_t))]}\}_N}$\\ \qquad\qquad\qquad\qquad\qquad\qquad\quad$+\overline{\lambda_e\{H(\pi_{\theta}(\cdot \mid S_t))\}_N}$

\STATE $L_{\omega}\leftarrow \overline{(\{R_t\}_N-R_m-V_\omega(\{S_t\}_N))^2}$

\STATE Minimize surrogate $L_{\theta},L_{\omega}$ w.r.t $\theta$ and $\omega$ with learning rate $\alpha_\theta, \alpha_\omega$
\ENDFOR
\end{algorithmic} 
\label{alg2}
\end{algorithm}

According to REINFORCE\cite{Williams1992,Sutton1999} and baseline techniques, Equation (\ref{con0}) can be optimized in the direction of,

\begin{equation}
\begin{split}
\nabla J(\theta|\bold S)&=\hat{A_t}*\nabla \ln P_{\theta}(A_t|S_t)
\\
&={(R_t-R_m-V_\omega(S_t))*\nabla \sum\limits_{i,j}\ln(\pi_{\theta}(A_t^{i,j}\mid S_t))}
\end{split}
\end{equation}

$\hat{A_t}$ denotes the advantage function estimated with $R_t-R_m-V_\omega(S_t)$, where $R_m$ indicates the moving average, and $V_\omega(S_t)$ is the value function estimated w.r.t. the encoding $\{enc\}$ by the critic module parameterized by $\omega$. $R_m$, updated by a batch of $R_t$ with rate $\alpha_{m}\in(0,1)$, is used to stabilize the training process by reducing the variance of the parametric baseline, which will be further discussed in Appendix. The critic is a 2-layer feed-forward network with ReLU layers in our experiment and trained to minimize the mean squared error $L_\omega$, between its value function estimate and the rewards with the moving average. 

It is shown in \cite{Williams1991} that an extra entropy regularization term can encourage the exploratory behaviour of the agent, and therefore we add it to the surrogate loss $L_\theta$  as,

\begin{equation}
H(\pi_{\theta}(\cdot \mid S_t))= -\frac{1}{n^2}\sum\limits_{i,j}[\pi_{\theta}^{i,j}(\cdot\mid S_t)* \ln(\pi_{\theta}^{i,j}(\cdot\mid S_t)]
\end{equation}

The REINFORCE with moving average baseline is given in Algorithm \ref{alg2}, where ${\{\cdot \}}_N$ means a batch of items and $\overline {\{\cdot \}}$ denotes an averaging operation for the batch. Through minimizing two surrogate losses $L_\theta$ and $L_\omega$, the actor and critic modules are trained using Adam optimizer with learning rates $\alpha_\theta, \alpha_\omega$ respectively.
\label{rein}

\subsection {Prioritized Sampling-Guided REINFORCE (PSR)}
\begin{algorithm}[H]
\caption{Prioritized Sampling-Guided REINFORCE}
\begin{algorithmic}[1]
\REQUIRE Replayer $R$ with size $S_R$; Batch size $N$; Moving average update rate $\alpha_{m}$; Learning rates for actor and critic $\alpha_\theta, \alpha_\omega$; Entropy term weight $\lambda_e$;
\FOR{$t=1,2,...$}
\STATE $\{\hat{A_t}\}_N\leftarrow \{R_t\}_N+R_m-V_\omega(\{S_t\}_N)$

\STATE Store $\{S_t, A_t, R_t,\hat{A_t}\}_N$ into $R$

\STATE Sample $\{S_t, A_t, R_t\}_N'$ from $R$ w.r.t Equation (\ref{pri})

\STATE $R_m\leftarrow (1-\alpha_{m})\overline{\{R_t'\}_N}+\alpha_{m}R_m$

\STATE $\{\hat{A_t'}\}_N\leftarrow \{R_t'\}_N+R_m-V_\omega(\{S_t'\}_N)$

\STATE $L_{\theta}\leftarrow \overline{\{{\hat{A_t'}*[\frac{1}{n^2}\sum\limits_{i,j}\ln(\pi_{\theta}(A_t'^{i,j}\mid S_t'))]}\}_N}$\\ \qquad\qquad\qquad\qquad\qquad\qquad\quad$+\overline{\lambda_e\{H(\pi_{\theta}(\cdot \mid S_t'))\}_N}$

\STATE $L_{\omega}\leftarrow \overline{(\{R_t'\}_N+R_m-V_\omega(\{S_t'\}_N))^2}$

\STATE Minimize surrogate $L_{\theta},L_{\omega}$ w.r.t $\theta$ and $\omega$ with learning rate $\alpha_\theta, \alpha_\omega$
\ENDFOR
\end{algorithmic} 
\label{alg3}
\end{algorithm}

We start to refine the original training method by incorporating prioritized sampling. The insight to adopt a replay buffer in this paper is to remember rare but latently useful experience for update and to help ensure the i.i.d assumption of estimates during update. On this basis, prioritized sampling offers a better strategy to select and reuse some useful experience batches more frequently, and therefore helps to learn more efficiently.

We measure the transitions replay frequency by the magnitude of their advantage function $\hat{A_t}$. In Schaul et al.\cite{Schaul2015}, two ways of prioritization are proposed for a certain transition $i$; proportional prioritization defines the priority $p_i$ directly according to the absolute value of $\hat{A_t}$ whilst rank-based prioritization defines $p_i$ as $\frac{1}{rank(i)}$ where the $rank(i)$ is sorted w.r.t. $|\hat{A_t}|$. As the rank-based prioritization is less prone to abnormal values of $|\hat{A_t}|$ and shows more robustness empirically, it is adopted in our paper\cite{Schaul2015}. 

To guarantee the discrete replay probability function as monotonic and non-zero, the probability $P(i)$ for transition $i$ being replayed is given as,
\begin{equation}
P(i)=\frac{p_i^\beta}{\sum_{i=1}^Np_i^\beta}=\frac{\frac{1}{rank(i)}^\beta}{\sum_{i=1}^N\frac{1}{rank(i)}^\beta}
\label{pri}
\end{equation}
 
Generally, experience replay introduces bias as it change the distribution of the estimates without necessary control and the solution when converged would also change as a result. In these cases, importance sampling ratio for each transition is required to adjust the optimization. However, in the setting of this paper, this bias remains trivial as each trajectory only has one step, and can be further abated with a comparatively small buffer size. 

With prioritized sampling, PSR shows faster convergence than the REINFORCE algorithm. However, as such improvement cannot change its optimization nature as a policy-gradient algorithm, its benefit is circumscribed. The major imperfection of such policy gradient-based method is that they are susceptible to severe policy degradation stemmed from an improper step size, which remain unsolved with the guidance of prioritized sampling.
\subsection {Trust Region-navigated Clipping Policy Optimization (TRC)}
TRPO addresses the aforementioned issue by imposing a trust region constraint on the objective function to limit the KL divergence between the old and new policies. Specifically, in our paper, the update direction and its constraint are given as,
\begin{equation}
\begin{split}
\nabla J(\theta|\bold S) =\hat{A_t}*&\nabla \frac{P_{\theta}(A_t|S_t)}{P_{b}(A_t|S_t)}={\hat{A_t}*\nabla \prod \limits_{i,j}\frac{\pi_{\theta}^{i,j}(A_t\mid S_t)}{b^{i,j}(A_t\mid S_t)}}
\\
& \bold{s.t.} \quad D_{KL}^{i,j}(b,\pi_\theta|A_t, S_t)< \sigma
\end{split}
\end{equation}

where $b(\cdot\mid S_t)$ represents the old policy. As noted in \ref{fspa}, each entry in $\pi_{\theta}(A_t\mid S_t)$ represents the probability for an  independent subaction whose action space is $\{0,1\}$, thus the Kullback-Leibler (KL) divergence $D_{KL}^{i,j}(b,\pi_\theta|A_t, S_t)$, between old and new policies given $A_t$ and $S_t$, is calculated as,

\begin{equation}
\begin{split}
D_{KL}^{i,j}(b,\pi_\theta|A_t, S_t)
=&b^{i,j}(A_t\mid S_t)\ln\frac{b^{i,j}(A_t\mid S_t)}{\pi_{\theta}^{i,j}(A_t\mid S_t)}+
\\
(1-&b^{i,j}(A_t\mid S_t))\ln\frac{1-b^{i,j}(A_t\mid S_t)}{1-\pi_{\theta}^{i,j}(A_t\mid S_t)}
\end{split}
\label{kl}
\end{equation}

\begin{algorithm}[htbp]
\caption{Trust Region-navigated Clipping Policy Optimization}
\begin{algorithmic}[1]
\REQUIRE Replayer $R$; Batch size $N$; Moving average update rate $\alpha_{m}$; Learning rates for actor and critic $\alpha_\theta, \alpha_\omega$; entropy term weight $\lambda_e$;
\FOR{$t=1,2,...$}
\STATE $\{\hat{A_t}\}_N\leftarrow \{R_t\}_N+R_m-V_\omega(\{S_t\}_N)$

\STATE Store $\{S_t, A_t, R_t, b(A_t\mid S_t),\hat{A_t}\}_N$ into $R$

\STATE Sample $\{S_t, A_t, R_t, b(A_t\mid S_t), \hat{A_t}\}_N'$ from $R$

\STATE $R_m\leftarrow (1-\alpha_{m})\overline{\{R_t'\}_N}+\alpha_{m}R_m$

\STATE Calculate ratio maps $\{ratio\}_N \leftarrow \{\frac{\pi_{\theta}(A_t\mid S_t)}{b(A_t\mid S_t)}\}_N$ 

\STATE Calculate KL-divergence maps $\{D_{KL}(b,\pi_\theta|A_t, S_t)\}_N$ w.r.t Equation (\ref{kl})

\STATE Clip ratio maps using KL-divergence maps w.r.t Equation (\ref{trc})

\STATE $L_{\theta}\leftarrow \overline{\{{\hat{A_t'}*\mathop{\Pi}\limits_{i,j}ratio^{i,j}}\}_N+\lambda_e\{ H(\pi_{\theta}(\cdot \mid S_t'))\}_N}$

\STATE $L_{\omega}\leftarrow \overline{(\{R_t'\}_N+R_m-V_\omega(\{S_t'\}_N))^2}$

\STATE Minimize surrogate $L_{\theta},L_{\omega}$ w.r.t $\theta$ and $\omega$ with learning rate $\alpha_\theta, \alpha_\omega$
\ENDFOR
\end{algorithmic} 
\label{alg4}
\end{algorithm}

This complicated second-order form makes the optimization process extremely computationally-expensive, particularly when extending to complex neural network.	Still, conjugate gradients method, which TRPO relies on, makes implementation complex and introduces more hyperparameters, the tuning of which becomes even more difficult when the algorithm requires heavy computation. PPO (generally the clipping version), being a first-order optimization method, vastly reduces the complexity by adopting a clipping mechanism to avoid dealing with the hard constraint. However, there exists an tiny but evident gap between the heuristic likelihood ratio constraint and trust region constraint\cite{Ilyas2018}. When we use PPO to calculate the joint ratio $\frac{P_{\theta}(A_t|S_t)}{P_{b}(A_t|S_t)}$, which is the product of $n*(n-1)$ individual ratios, the deviation between the two constraints for each ratio would result in horrendous aggregate deviation after product calculation during optimization. On the other hand, the adaptive KL penalty version\cite{Schulman2017} of PPO is also computationally-inefficient as the optimization objective involves a penalty term for KL divergence, which is a source of huge computation during the back-propagation process. 

Given these problems, we try to adopt a better clipping strategy to adopt the efficiency of PPO whilst abating the constraints gap. Inspired by the mechanism that PPO employs the likelihood ratio as the triggering condition for clipping, we replace ratio-based triggering condition with a trust-region one. The likelihood ratio is clipped only when the policy is out of the trust region. Specifically, 
\begin{eqnarray}
ratio^{i,j}=
\begin{cases}
clip(ratio^{i,j}, 1-\epsilon, 1+\epsilon) & D_{KL}^{i,j}\geq\sigma\\
ratio^{i,j} & D_{KL}^{i,j}< \sigma
\end{cases}
\label{trc}
\end{eqnarray}

In our implementation, the clipped ratio maps are stored temporarily after clipping, and given the KL-divergence maps, the TRC ratio maps are subsequently generated by selecting clipped ratios or non-clipped ones w.r.t Equation (\ref{trc}). In this sense, the behavioural difference between PPO and TRC can be studied according to these maps (See discussions in section \ref{exp1}). TRC utilizes the same prioritized sampling strategy of PSR.

\subsection {Reinforcement Learning Procedure}

This section gives the reinforcement learning procedure within which the three aforementioned algorithms is adopted in turns to optimize the search efficiency of RL agent. It is proved by Zhu et al\cite{Zhu2020}, through maximizing the reward over all directed graph space, the DAG with the best scoring is obtainable on condition that
\begin{equation}
\lambda_1 \min_{A_t\not\in DAGs}{CL(A_t)}+\lambda_2 \geq {BIC}_u-{BIC}_l\label{con1}
\end{equation}

where the upper bound ${BIC}_u$ is easily calculated by some random DAGs whilst the lower bound ${BIC}_l$ can be simply set as zero with the independence-based score\cite{Peters2014}. Even though $\min_{A_t\not\in DAGs}{CL(A_t)}$ is hard to compute, by setting $\lambda_2={BIC}_u-{BIC}_l$, Constraint (\ref{con1}) is assured for any $\lambda_1\geq0$. We set a relatively small value for $\lambda_1$ to facilitate accurate DAG generation. 
The overall reinforcement learning procedure is given in Algorithm \ref {algorithm1}.

\begin{algorithm}[htbp]
\caption{Reinforcement Learning Procedure}
\label{algorithm1}
\begin{algorithmic}[1] 
\REQUIRE Score Parameters: $BIC_u, BIC_l, BIC_0$; Penality Weights: $\lambda_1, \Delta_1, \Lambda_1$ and $\lambda_2, \Delta_2$; Parameters Update Iteration: $t_u$; 
\FOR{$t=1,2,...$} 
\STATE Sample $S_t$ from $X$, and generate $A_t$ with $\pi(A_t \mid S_t)$
\STATE $BIC(A_t)\leftarrow BIC_0*\frac{(BIC(A_t)-BIC_l)}{(BIC_u-BIC_l)}$
\STATE $R_t= -[BIC(A_t )+\lambda_1 CL(A_t )+\lambda_2 Ind(A_t )]$
\STATE Optimize $\pi(A_t \mid S_t)$ with Algorithm \ref{alg2}, \ref{alg3} or \ref{alg4}
 \IF{$t \mod t_u = 0$}
  \IF {the best-rewarded graph is a DAG with $BIC_{min}$}
  \STATE $BIC_u\leftarrow \min(BIC_u, BIC_{min})$
  \ENDIF
 \STATE $\lambda_1\leftarrow \min(\lambda_1\Delta_1, \Lambda_1)$, $\lambda_2\leftarrow \min(\lambda_2+\Delta_2, BIC_u)$
 \STATE Update recorded rewards w.r.t $\lambda_1, \lambda_2$
 \ENDIF
\ENDFOR
\end{algorithmic} 
\end{algorithm}
Similar to Lagrangian methods adopted in NOTEARS\cite{Zheng2018}, to ensure Constraint (\ref{con1}), $\lambda_1$ and $\lambda_2$ start with small values and are gradually increased, incrementally by $\Delta_1$ with upper bound $\Lambda_1$ and multiplicatively by $\Delta_2$ with upper bound $BIC_u$ respectively. Meanwhile,  as the score function is unbounded whilst $CL(A_t)$ and $Ind(A_t)$ are independent of the range of the score function, the predefined score is restrained to $[0, BIC_0]$ with $BIC_0*\frac{(BIC(A_t)-BIC_l)}{(BIC_u-BIC_l)}$. $BIC_l$ is calculated with a complete directed graph, whereas $BIC_u$ is computed with an empty graph initially and is further adjusted with the recorded lowest score $BIC_{min}$ during DAG generation periodically. 
\section {Experiments}
\label{sec6}
\noindent As we wish to obtain the best-rewarded graph, all the graphs generated during training are recorded. Before we output the best-rewarded graph, a pruning process is further needed as it may entail spurious edges. For a discovered causality relation, we iteratively remove a parental variable once without changing other causal relations and the new graph would be accepted if the performance of the resulting graph does not decline or decline within a predefined acceptable range, or be refused otherwise.

We compare our method against RL-BIC/RL-BIC2\cite{Zhu2020} along with other traditional or recent gradient-based approaches on synthetic and benchmark real dataset. These methods include PC\cite{Spirtes2000}, GES (with BIC)\cite{Ramsey2017}, ICA-LiNGAM\cite{Shimizu2006}, Causal Additive Model (CAM)\cite{Buhlmann2013}, NOTEARS\cite{Zheng2018}, GraN-DAG\cite{Lachapelle2019} and DAG-GNN\cite{Yu2019}. Implementations of these algorithms would follow the default hyperparameters in their papers unless stated otherwise. For pruning, the same threshold method for NOTEARS and ICA-LiNGAM is adopted. RL-BIC/RL-BIC2 denotes the REINFORCE algorithm specified in \ref{rein} using the two versions of BIC score given in Equation (\ref{bic1}) and (\ref{bic2}). We likewise denote our proposed algorithms using the two versions of BIC as PSR-BIC/PSR-BIC2 and TRC-BIC/TRC-BIC2 accordingly.

The empirical results are examined with following metrics: structural hamming distance (SHD), false discovery rate (FDR) and true positive rate (TPR) w.r.t the true graph. SHD, the main indicator, is the minimum operation needed to revert the estimated graph to the true graph. Lower SHD indicates better estimate of the causal graph. For the fair comparison among the RL algorithms, all the causality model settings and RL procedural parameters would be specified and remain the same in each of the experiments. (see Appendix for more details)

\subsection {Linear-Gaussian and LiNGAM}
\begin{figure*}[htbp]
  \center
  \includegraphics [width=18cm,height=4.5cm]{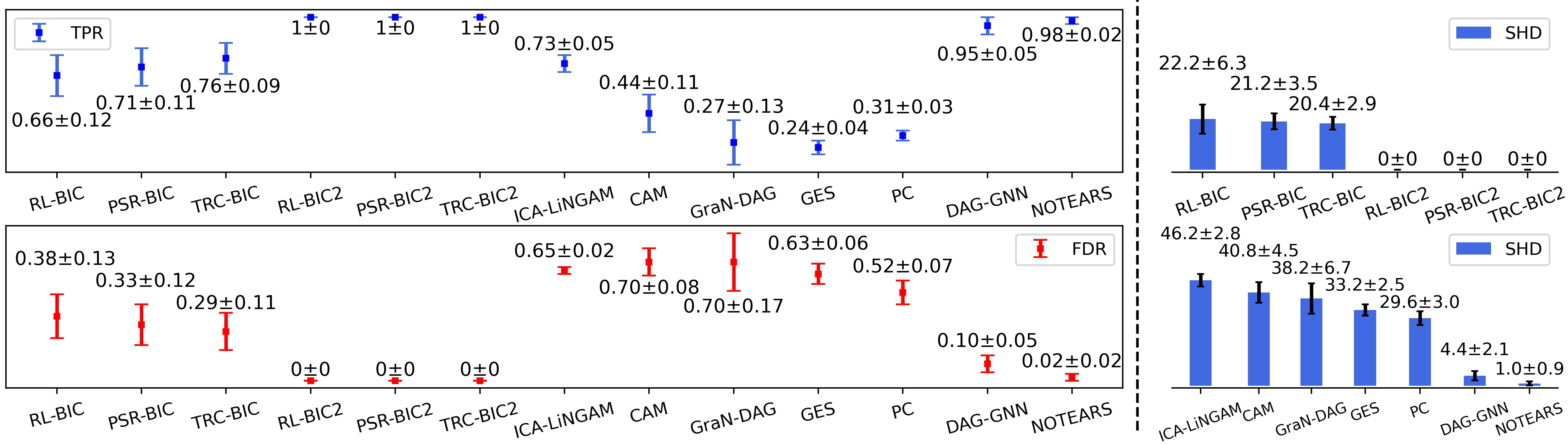}
  \caption{Empirical Results on Linear-Gaussian Data}
  \label{fig4}
\end{figure*}

\begin{figure*}[htbp]
  \center
  \includegraphics [width=18cm,height=4.5cm]{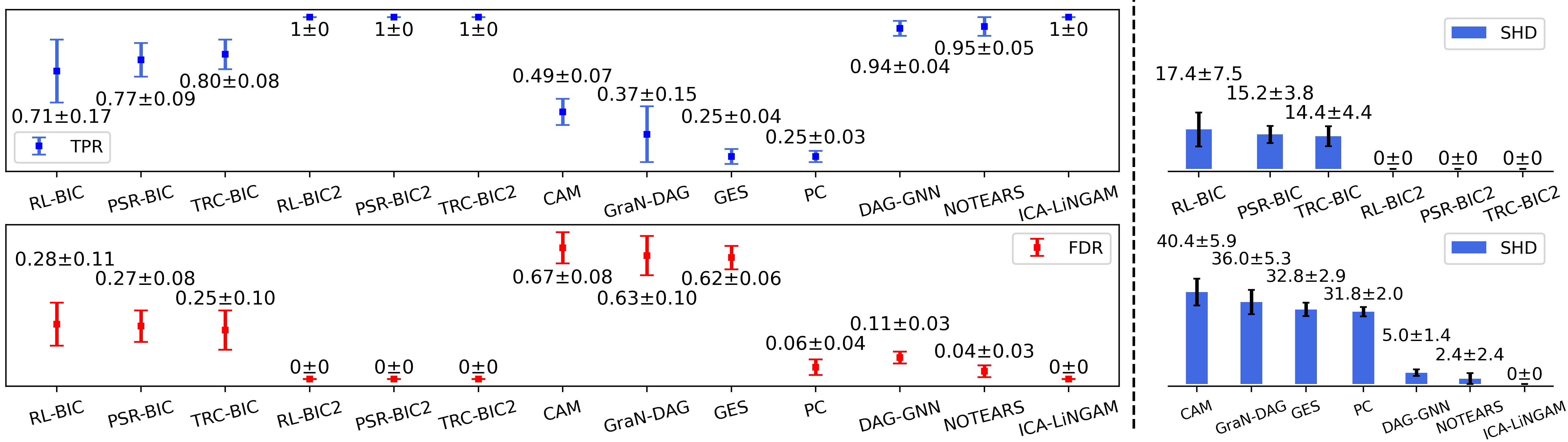}
  \caption{Empirical Results on LiNGAM Data}
  \label{fig5}
\end{figure*}

For an $n$-variable causality model, an $n*n$ upper triangular matrix is sampled as the adjacency matrix with upper elements individually drawn from Bernoulli$(0.5)$. The elements in edge weight matrix $W \in \mathbbm {R}^{n\times n}$are sampled independently from Uniform$([-2,-0.5]\cup[0.5,2])$ to generate the node features w.r.t. Equation (\ref{gen1}) from both noise models. All noises have unit variances and the non-Gaussian noise is obtained by passing Gaussian-sampled disturbance variables to a power non-linearity\cite{Shimizu2006}. The pruning process is done by thresholding the estimated coefficients with rate 0.3. The results on linear-Gaussian and LiNGAM with $n=12$ nodes and 5000 samples are given in Fig. \ref{fig4} and \ref{fig5}. For each data model, the results for RL-based algorithms and non-RL-based algorithms are separated and both ranked by their SHD. 

\begin{figure*}[htbp]
\centering
\subfigure[$\lambda_1$]{
\label{fig9123:a} 
\includegraphics[width=4cm,height=6cm]{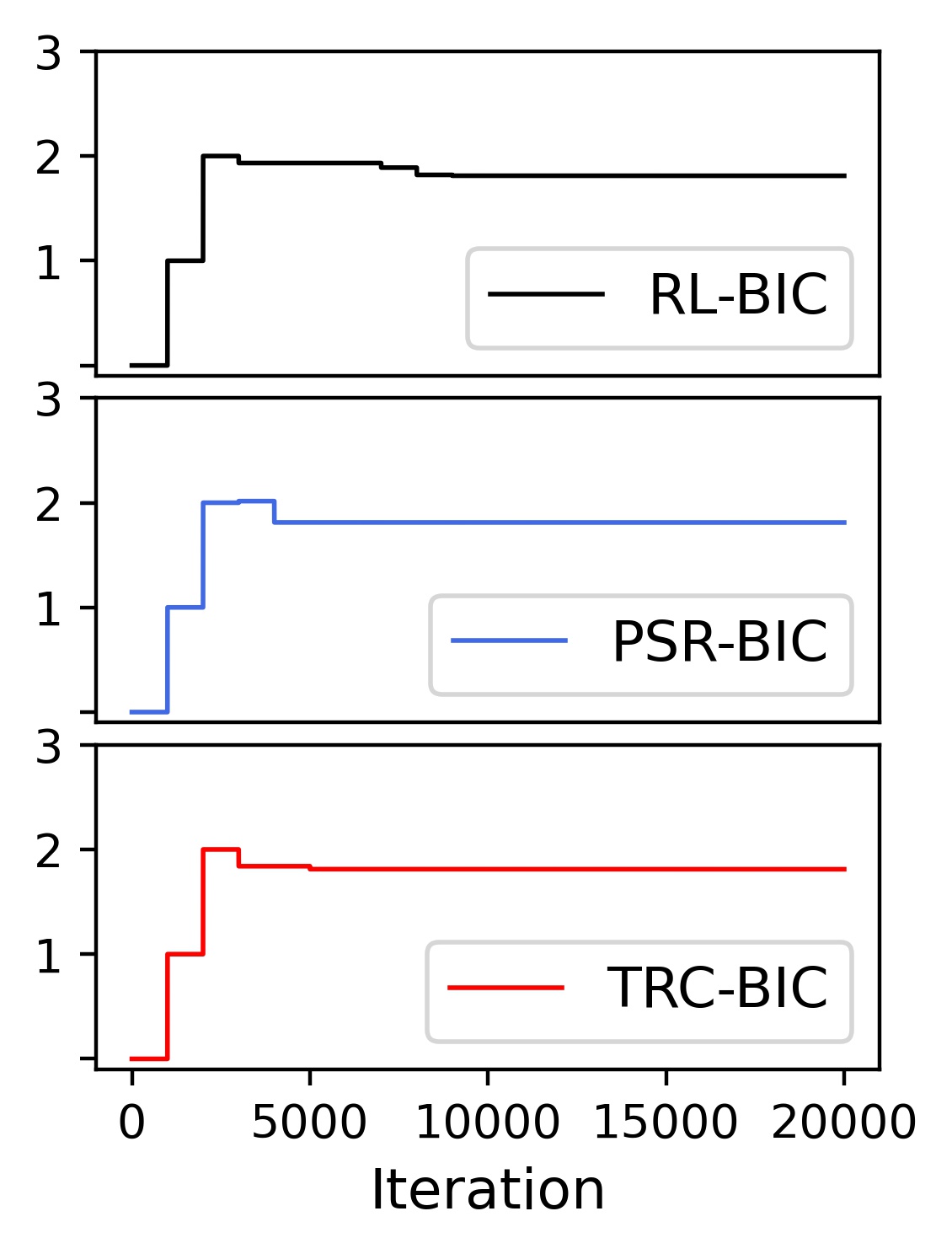}}
\hspace{0.02in}
\subfigure[$\lambda_2$]{
\label{fig9123:b} 
\includegraphics[width=5cm,height=6cm]{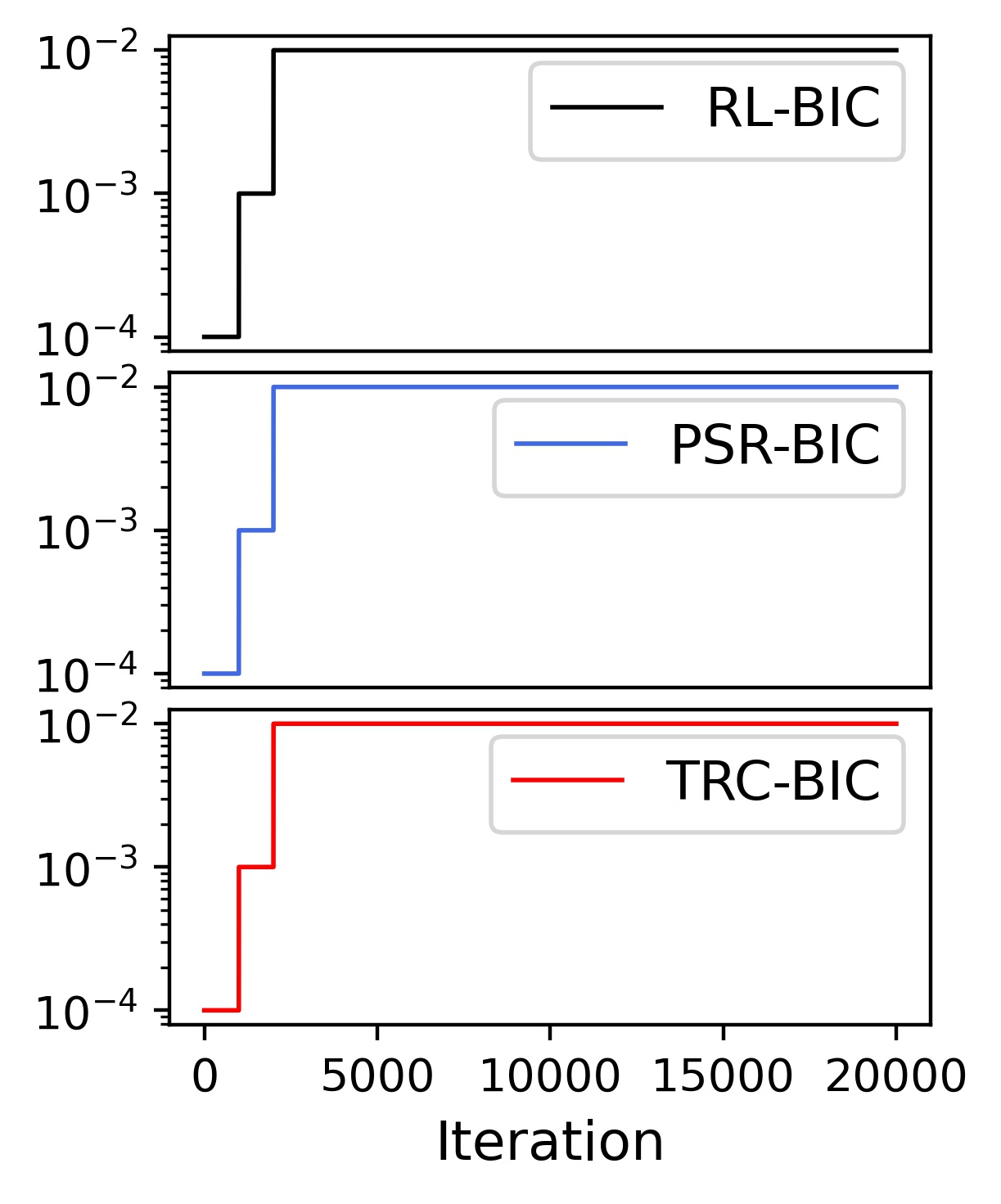}}
\hspace{0.02in}
\subfigure[Batch Negative Reward]{
\label{fig9123:c} 
\includegraphics[width=8cm,height=6cm]{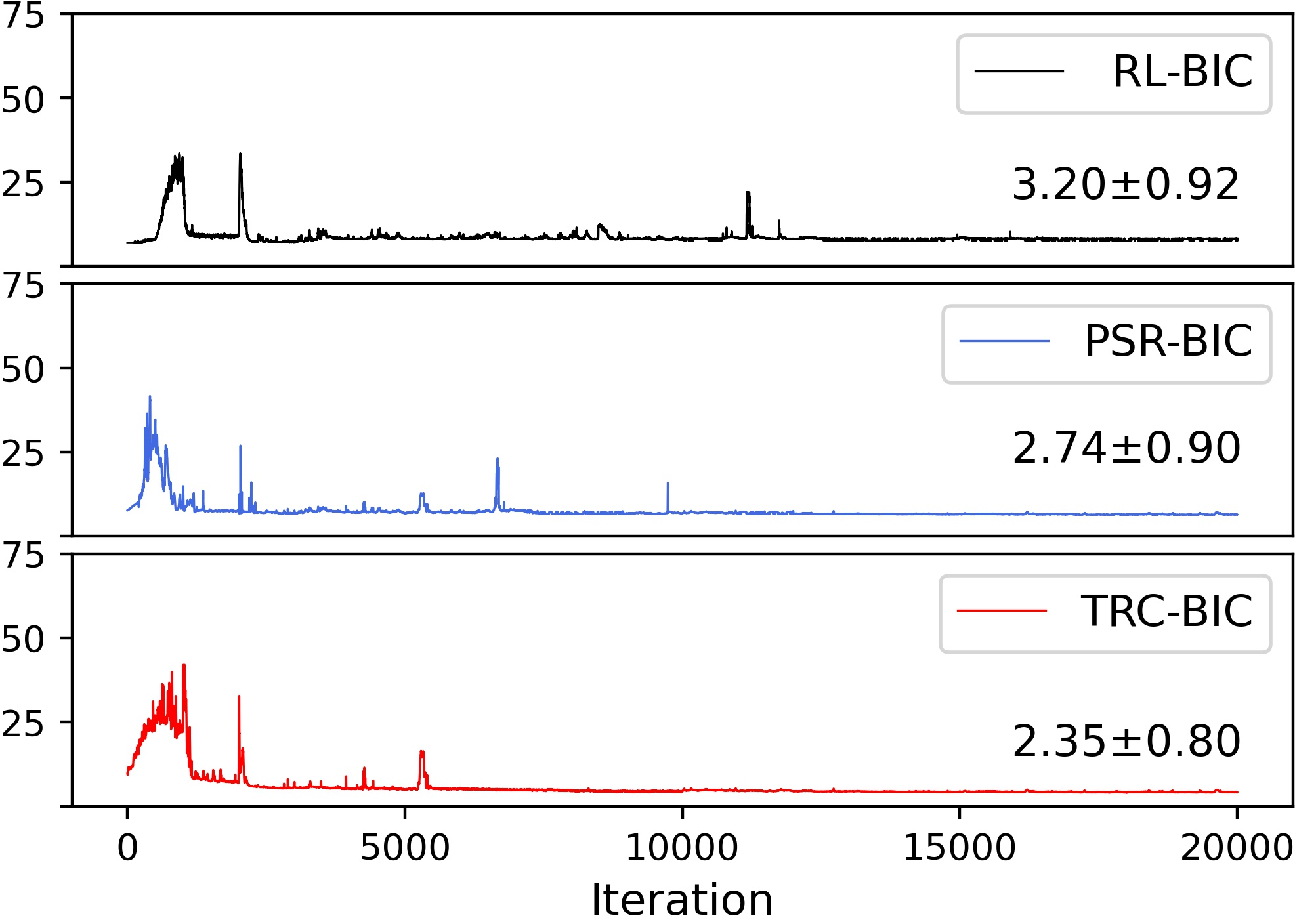}}
\caption{Comparison of the Learning Process using Different RL Algorithms}
\label{fig9123} 
\end{figure*}

For non-RL-based methods, DAG-GNN and NOTEARS produce favorable causal graph results on both models thanks to the decent perception ability of their carefully-crafted neural network designs. ICA-LiNGAM could not guarantee performance in linear-Gaussian models but projects competence in LiNGAM owing to its major assumption on data models. The performance of GraN-DAG on linear models is chained partly because its two-layer feed-forward network design could not capture linear relations. CAM is mainly used on nonlinear models for its assumption and shows miserable results on linear models.

We observe that for both linear models, all RL-based approaches adopting $BIC_2$ recovers all the true causal graphs whereas those with $BIC_1$ is comparatively under-performed. We remark that $BIC_2$ is the better reward signal barely in this linear case, and as we shall see in following experiments, not always sufficient due to its noise assumption. On the other hand, with the improved robustness and search efficiency of both PSR and TRC, PSR-BIC and TRC-BIC outperforms RL-BIC on both datasets in terms of the generated DAG accuracy.

\begin{figure*}[htbp]
\centering
\subfigure[Different Clipping Bounds]{
\label{figa_b:a} 
\includegraphics[width=5cm,height=5cm]{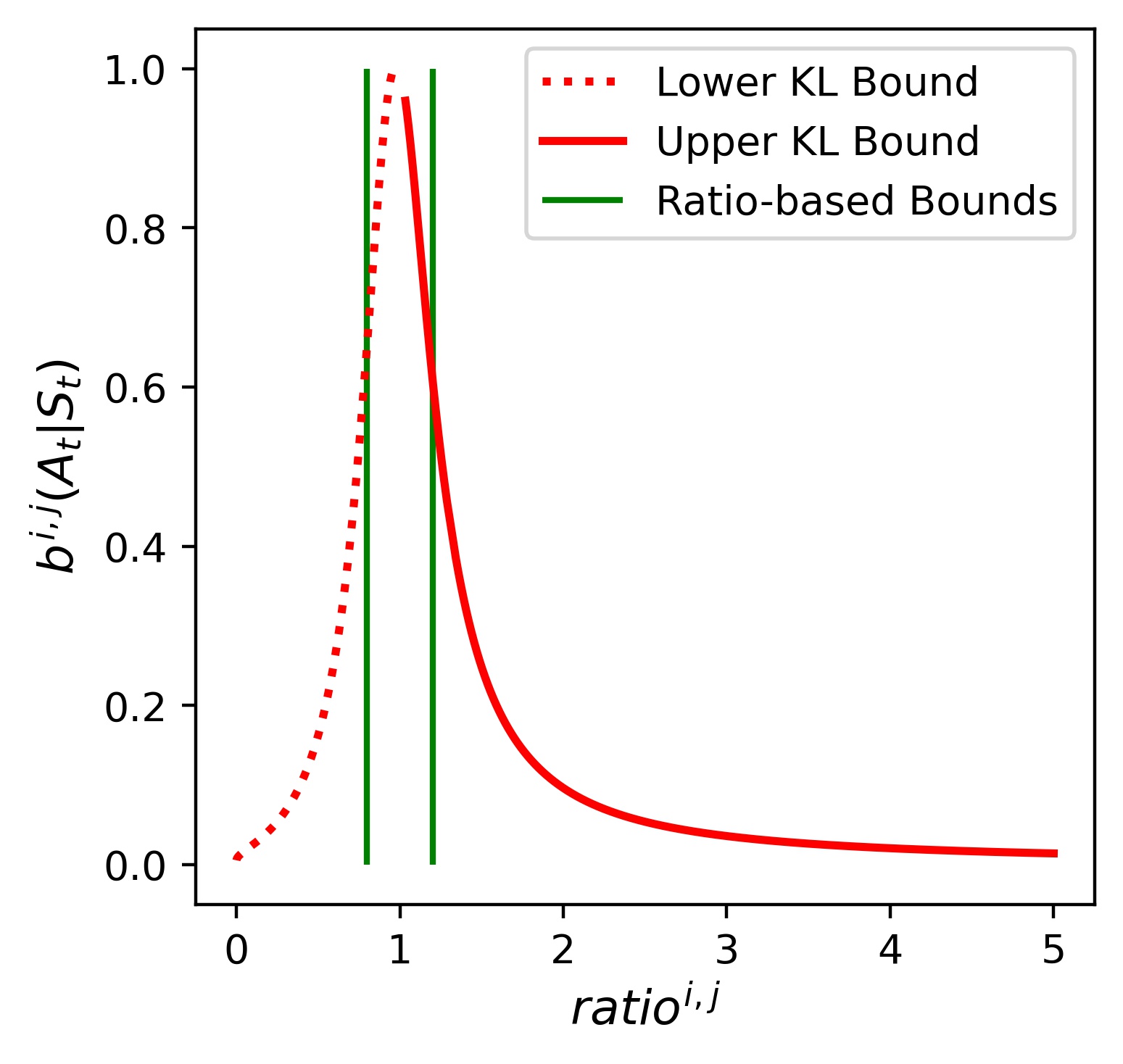}}
\hspace{0.05in}
\subfigure[Probability of being Clipped (\%) for PPO (Left) and TRC (Right)]{
\label{figa_b:b} 
\includegraphics[width=10cm,height=5cm]{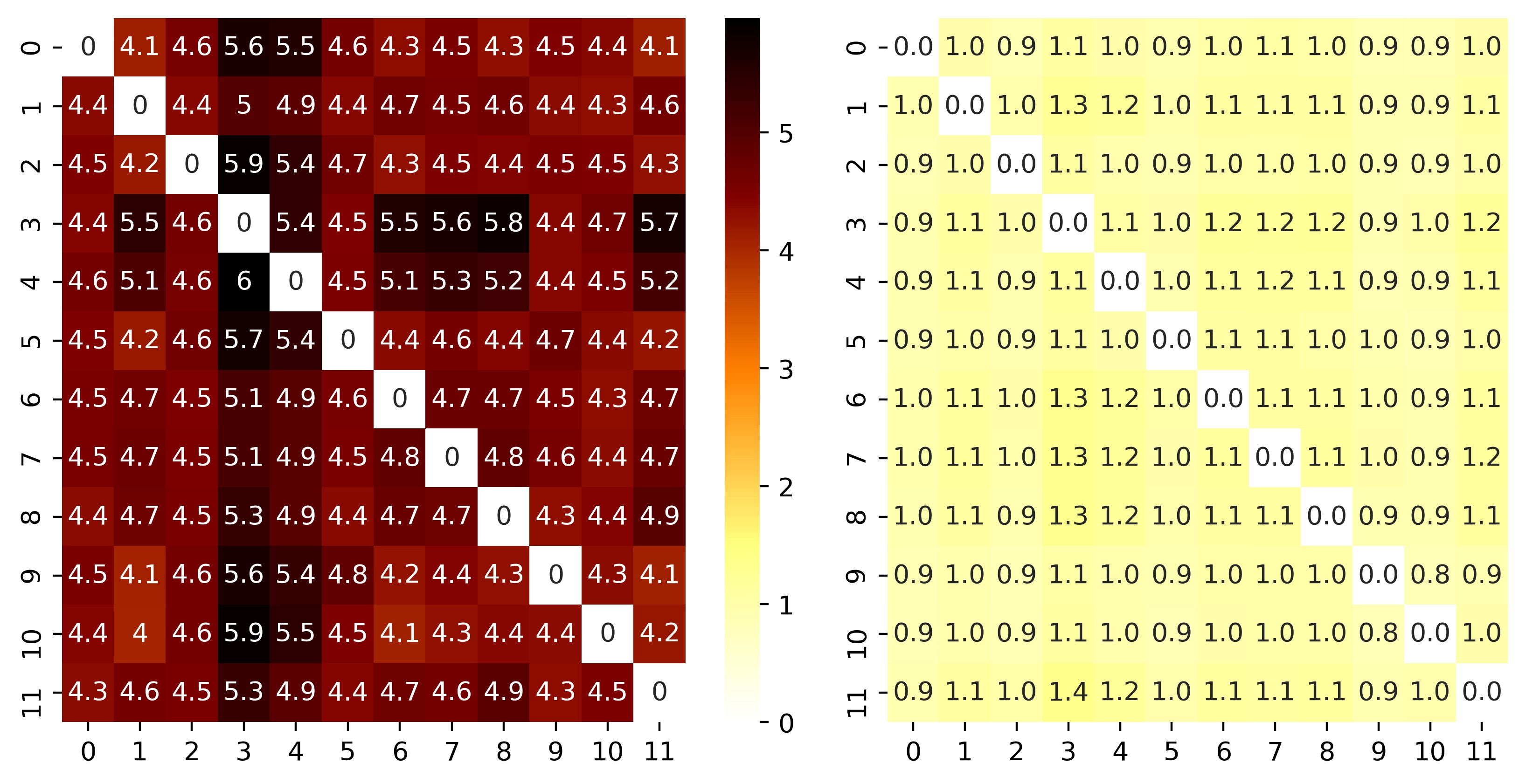}
}
\caption{Behavioural Difference of TRC and PPO}
\label{figa_b} 
\end{figure*}

We compare the learning behaviour of the three RL algorithms on linear models to test their search efficiency using $BIC_1$. Similar behavioural difference is observed when handling nonlinear models and real-world data. We notice that, as shown in Fig.  \ref{fig9123:c}, REINFORCE is much slower to converge both in the short-term and long-term. In the short term, i.e. within each training period before the hyperparameters update, REINFORCE manifests disappointingly poor search efficiency and slow improvement in negative reward before the first hyperparameters update or once abnormality is detected. Interestingly, the curves for TRC and PSR show comparatively larger jitters before the first hyperparameters update but would plummet promptly. This is because the prioritized replay they both adopted make them aware of the abnormal reward signal and react immediately and attentively. In the long term, as indicated by the mean and standard deviation of rewards for the last 5000 iterations shown above the curve tail in Fig. \ref{fig9123:c}, TRC would stop fluctuating or fluctuate within a tolerated range much earlier than REINFORCE and PSR, and coverge at the best point (-2.35). This phenomenon could also be observed through the value of $\lambda_1$, as shown in Fig. \ref{fig9123:a}, where $\lambda_1$ is subject to much slower adjustment by the RL procedure with REINFORCE. $\lambda_2$, on the other hand, does not project such difference probably because it changes multiplicatively each time when needed and therefore less subject to adjustment.

We also examined the behaviour difference of TRC and PPO under linear settings. As can be concluded in Fig. \ref{figa_b:a}, where the clipping range for TRC and PPO under different value of  $b^{i,j}(A_t|S_t)$ are shown, PPO's metric with a constant clipping range imposes a stricter constraint than the trust region constraint on subactions that are not preferred by the old policy, i.e. when the value of a certain subpolicy $b^{i,j}(A_t|S_t)$ remains small. This would lead to an exploration issue particularly when the decision-making process has to cope with considerable subaction simultaneously. TRC relax such constraint by widening the gap between the upper and lower clipping bounds.

Fig. \ref{figa_b:b} gives an example of the overall probability of being clipped for each ratio with a 12-node setting in Experiment \ref{exp1} using TRC and PPO during the whole training process. We observe that using PPO, each ratio is roughly 4 to 5 times easier to be  subject to clipping than using TRC. With such superfluous clipping times, PPO would strictly restrain exploratory behaviour of the agent on the potential valuable subactions that are not preferred by the initial policy. Through better clipping strategy, TRC manifests more exploratory attempts whilst staying safe in the KL bounds. The output results for PPO turn out to be miserable in terms of the three metrics, with SHD in both scenarios over 35, and are therefore not included in our comparative studies.

\label{exp1}

\subsection {Non-linear Models with Quadratic Functions or Gaussian Processes}

\begin{figure*}[htbp]
  \center
  \includegraphics [width=18cm,height=4.5cm]{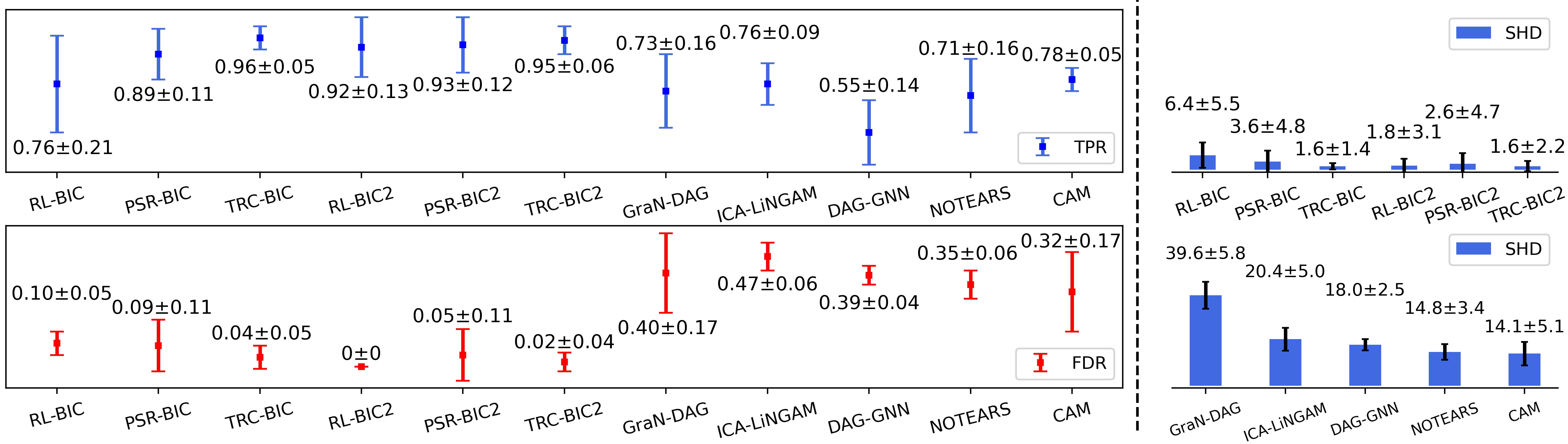}
  \caption{Empirical Results on Non-linear Models with Quadratic Functions}
  \label{fig8}
\end{figure*}

\begin{figure*}[htbp]
  \center
  \includegraphics [width=18cm,height=4.5cm]{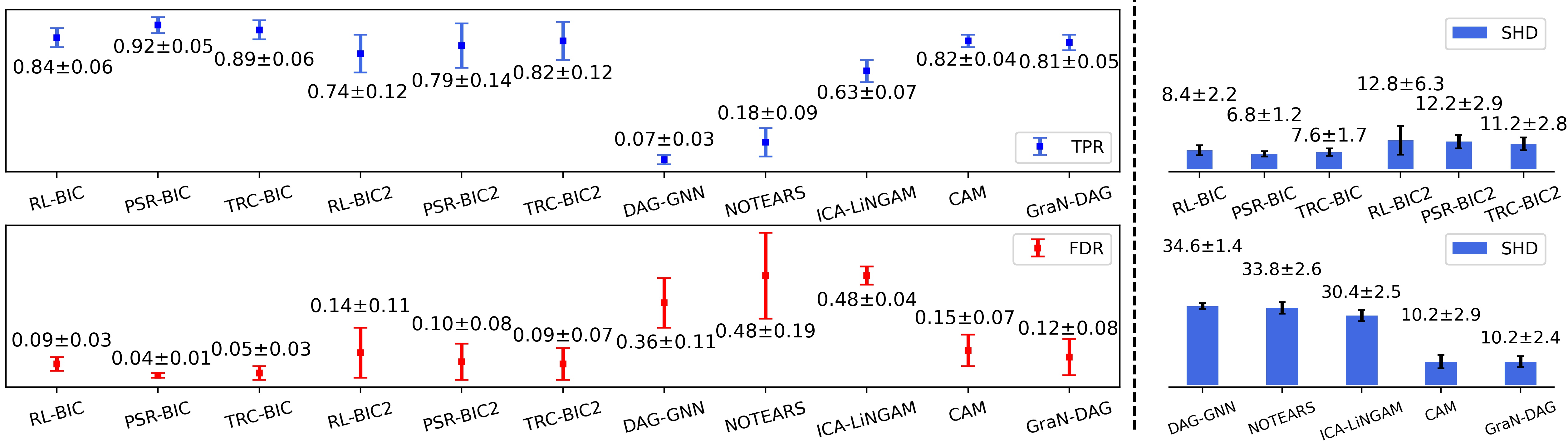}
  \caption{Empirical Results on Non-linear Models with GP}
  \label{fig9}
\end{figure*}

For both non-linear models, the upper triangular matrix is sampled in a similar way discussed above. Concerning the quadratic model, given the parents set implied in the upper triangular matrix for the $i$-th node, a function combing all the first-order and second-order features is generated with the coefficients sampled from Uniform$([-1,-0.5]\cup[0.5,1])$ or being zero at same rate. With the non-Gaussian noises setting, a 10-node dataset with 5000 samples is generated, and we employ quadratic regression to model causality. The pruning process is done by thresholding all the coefficients for both first-order and second-order terms. As for the GP model, each function describing the causal relations are generated from a GP with radial basis function of bandwidth 1 and the noise is drawn from normal distribution with uniformly-sampled variance. Therefore, GP Regression with RBF kernel is employed here to model the causal relations. The variable data are normalized and median heuristics are applied for kernel bandwidth to avoid over-fitting caused by fixed kernel bandwidth.

We observe that all RL-based approaches outperform the non-RL-based ones in the two non-linear scenarios. In the quadratic scenario, although REINFORCE could already achieve decent results in terms of the three metrics, yet TRC could further improve the output performance and produce nearly the true graph (SHD$\leq 1$). Surprisingly, for the GP model, PSR would deliver the best result when combined with BIC, whilst REINFORCE still lag behind the other two RL approaches.
\label{nonlinearexp}

\subsection {Pseudo-Real Data and Real Data}

\begin{figure*}[htb]
  \center
  \includegraphics [width=18cm,height=5.625cm]{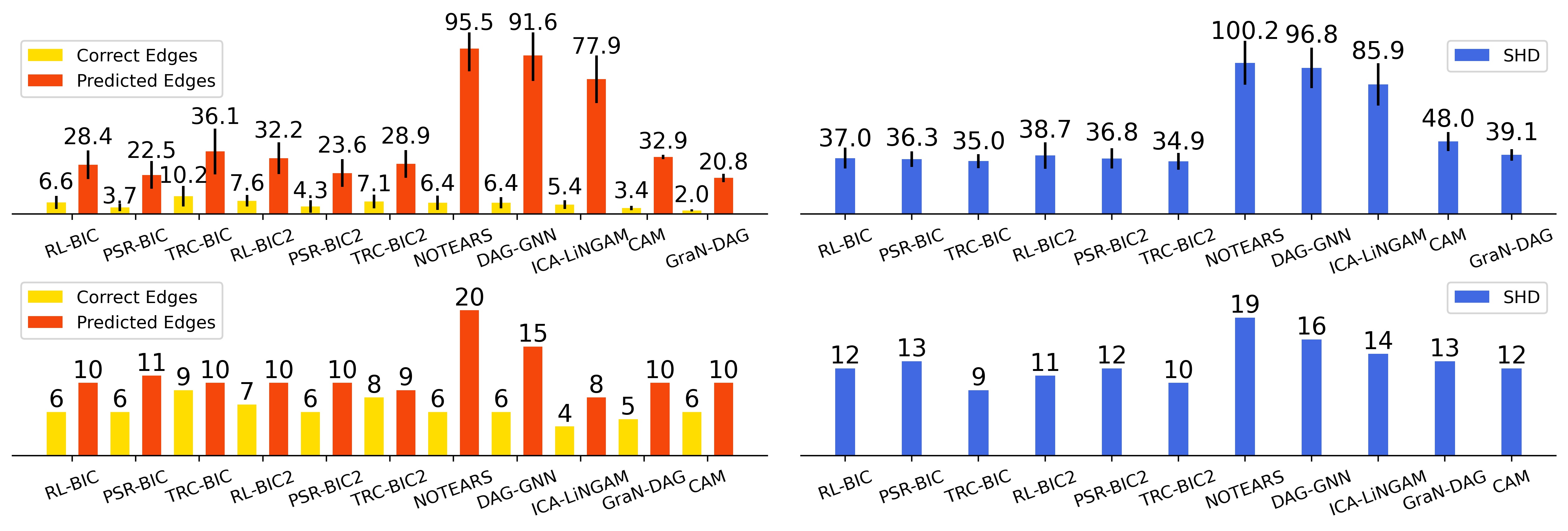}
  \caption{Empirical Results on SynTReN (above, 20 nodes) and CYTO (below, 14 nodes)}
  \label{fig10}
\end{figure*}

\begin{figure*}[htbp]
\centering
\subfigure[Discovered Graph]{
\label{fig11:a} 
\includegraphics[width=8cm, height=4cm]{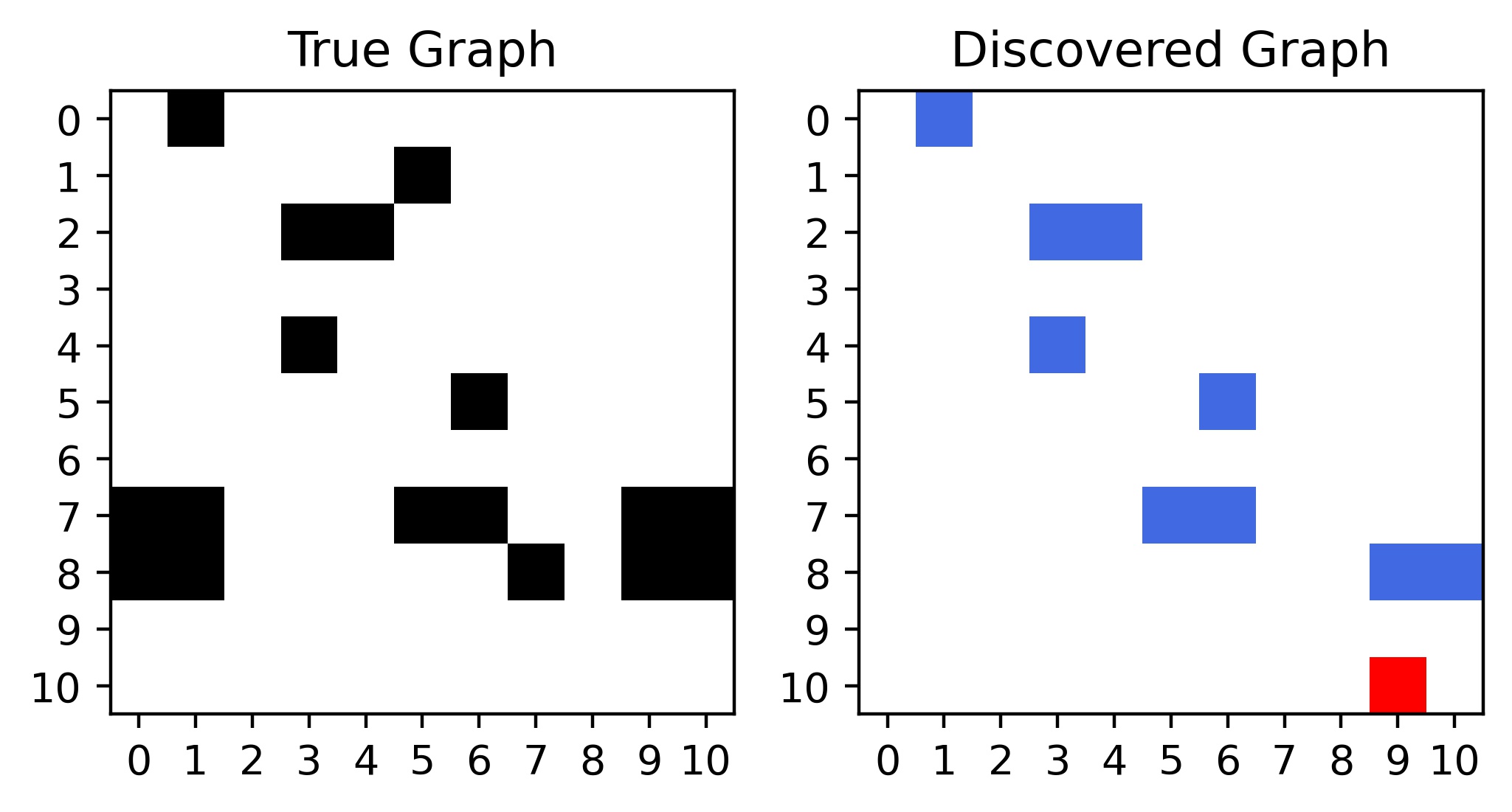}}
\hspace{0.4in}
\subfigure[Recovered Graph]{
\label{fig11:b} 
\includegraphics[width=6cm, height=4cm]{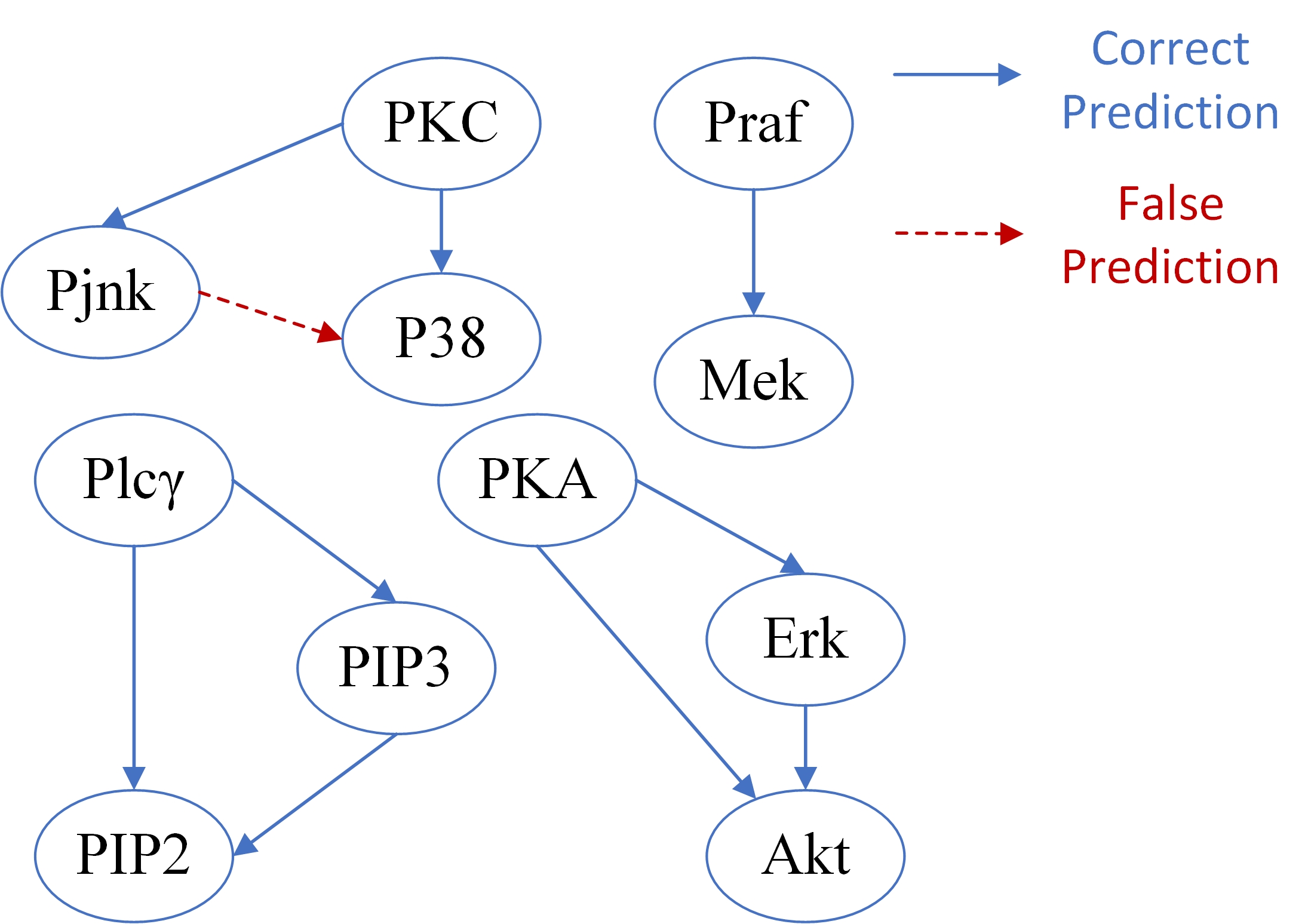}}
\caption{Edge Discovery Results on CYTO}
\label{fig11} 
\end{figure*}

\begin{figure*}[htbp]
  \center
  \includegraphics [width=18cm,height=4.5cm]{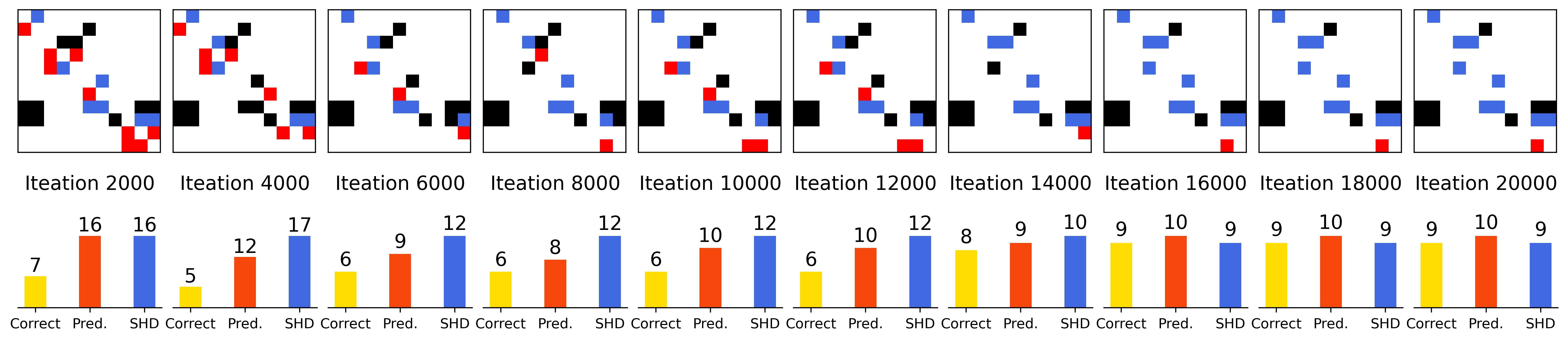}
  \caption{Evaluation of Results over Iterations on CYTO}
  \label{fig12}
\end{figure*}

We consider CYTO\cite{Sachs2005}, a real-world dataset consists of exactly 853 single cell recordings of the abundance of 11 phosphoproteins and phospholipids. As a benchmark for causality problems, it contains both observational and interventional data for a graph that has 11 nodes and 17 edges. We only use the observational data for the discovery task. As for pseudo-real data, we test our methods on ten datasets generated by the SynTReN \cite{BulckeLNRMVMM06} generator. The SynTReN datasets are simulated gene expression data that approximates real data after creation of networks representing transcriptional regulation. The true graph for SynTReN has 20 nodes and 24 edges. We follow the same normalization, regression and heuristics methods for the them as in the GP model in \ref{nonlinearexp} except that we use CAM pruning. 

We change the metrics in reporting the results on pseudo-real and real data as the two true causal graphs are sparse by its nature, which is the general phenomenon for real-world causality, and we aim to report metrics that contain more details concerning the estimated graph. The bar graphs of Fig. \ref{fig10} show the number of correct edges and predicted edges and SHD. For SynTReN, the results are averaged over the ten generated datasets. TRC-BIC and TRC-BIC2 surpass other methods on SynTReN by reaching the SHD of 35.0 and 34.9 respectively. They also achieve the hitherto best result on the protein dataset with the lowest SHD of 9 and 10 respectively. Fig. \ref{fig11} shows the discovered graph and its pertinent recovered graph that encodes the protein interactions. TRC-BIC helps detect 9 correct edges out of 10 discovered edges on CYTO, and 10.2 correct edges on SynTReN, outperforming all other methods.

We evaluate the generated graph with the true graph on CYTO every 2000 iterations, and figure out that the graphs are optimized in terms of the prediction accuracy and SHD until the best result is obtained, which manifest the convergence of the algorithm. For the discovered graph in Fig. \ref{fig11} and \ref{fig12}, correct prediction, false prediction and true edges are indicated by blue, red and black blocks respectively. Judging from the graph evolving process shown in Fig. \ref{fig12}, we observe that agent trained by the algorithm prefers to reverse relation for either false predictions or true predictions and afterwards listen closely to the reward signal. Once the negative signal flows back, it would withdraw the decision and sometimes abandon both relations once it finds out neither of them is appropriate. Guided by the reward signal, our algorithm prefer to select compact graphs with relatively few edges compared with other traditional methods.

\begin{figure}[!htbp]
\centering
\includegraphics[width=9cm,height=5.5cm]{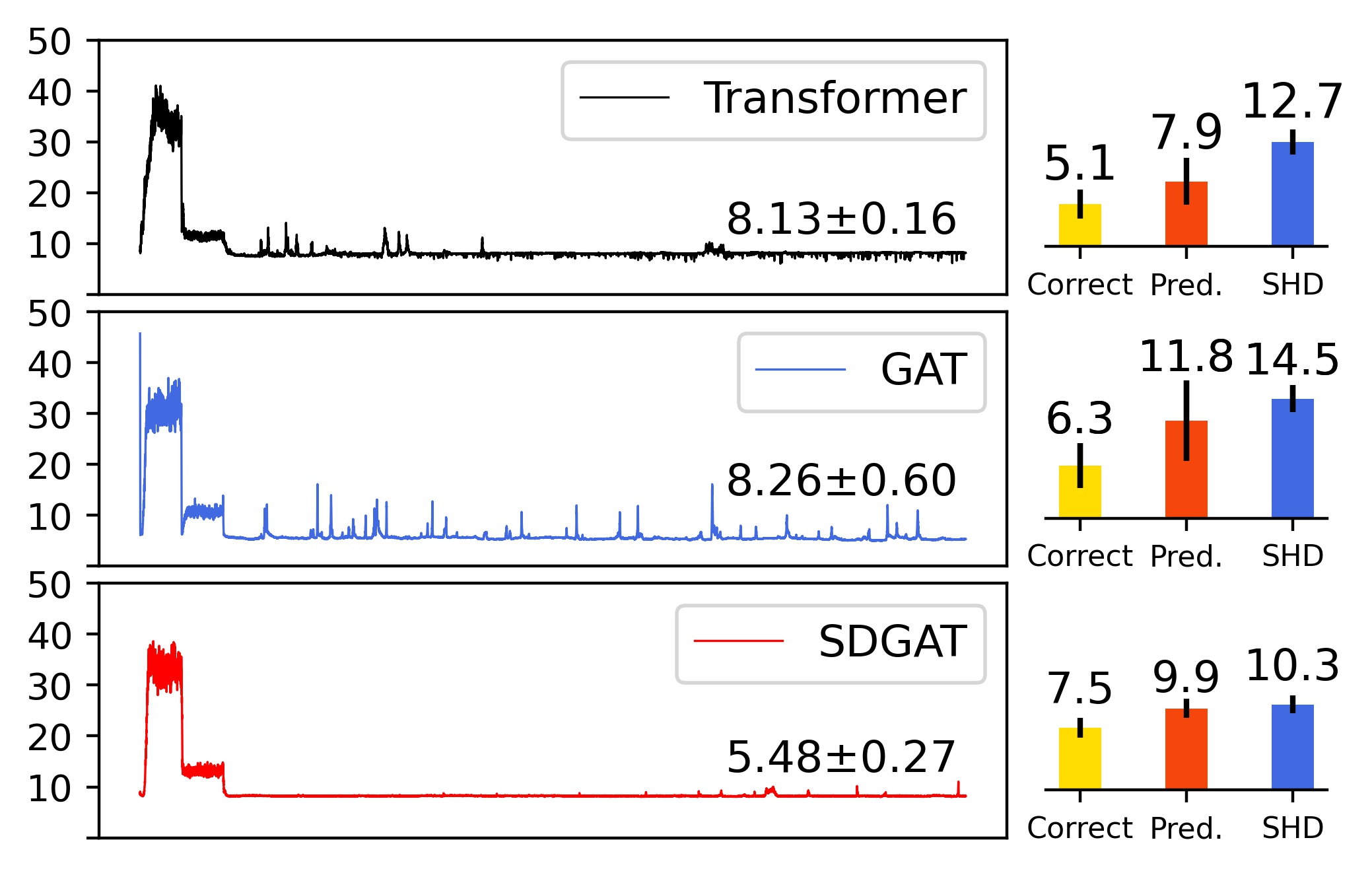}
\caption{Results using Different Encoders on CYTO}
\label{fig13} 
\end{figure}

We compare among the performance of different encoders under this setting to examine their ability to encode real-world causality. As the scaled dot-product attention mechanism is adopted in Transformer as well, we would add it to the comparative studies. We conduct three experiments using TRC-BIC on the CYTO dataset with only the encoders changed. Fig. \ref{fig13} shows the curves for rewards (left) and the three metrics for outputted graph (right) averaged over 20 training times. We observe that the SDGAT comes out the best encoder with TRC not only because the reward curve converges at -5.48 with standard deviation of 0.27 for the last 5000 iterations, but also it obtains the best results on average. Transformer also manifests decent search efficiency as it finds out 5.1 correct edges out of 7.9 predictions and outputs graphs of SHD 12.7. GAT performs the worst (with average SHD being 14.5) and fails to converge even after 15000 iterations. Compared with Transformer, SDGAT brings steadier convergence and better search efficiency in this RL task. The complexity analysis is given in TABLE \ref{tab1}.

\begin{table} [htbp]
\small
\caption{Network Complexity Analysis}  
\begin{center}  
\begin{tabular}{cc}  
\toprule[2pt]
\textbf{Architecture} & \textbf{Complexity per Layer}\\ \hline  
GAT & $O(nmm'+n^2m')$ \\ \hline  
Transformer& $O(n^2m)$\\  \hline  
SDGAT & $O(n^2m')$\\  \hline 
\end{tabular}  
\end{center}
\label{tab1}
\end{table}
\section {Conclusion}
\label{sec7}
\noindent We propose TRC to search for the best-rewarded DAG, bringing the hitherto best results and convergence speed in solving the RL task for causal discovery. We integrate prioritized sampling, as an optimization trick, into the original REINFORCE algorithm adopted in the original RL task for causal discovery. We also propose the SDGAT by upgrading the original GAT with the scaled dot-product attention. The effectiveness of the proposed methods and network designs is proved on both synthetic and benchmark real datasets.

However, there still exist great challenge when handling large graphs but small-sized datasets even for TRC because of the limitation of the current RL formulation. This issue mainly stems from the complexity of the score calculation, which depends on regression, and the inability to deploy such calculation on GPUs for parallel computing. Secondly, with SDGAT, it is also possible to inject priori knowledge to boost the DAG generation by reducing the search space, which is not considered in our current work. 

The proposed approach possesses decent scalability as each of its component can be refined individually to guide overall improvement. For example, given the time-consuming condition in the calculation of the scores, we would expect to use a neural network to estimate the score function. Still, the extraction and incorporation of priori knowledge would be promising to support the discovery of causality. Besides, Wang \cite{WangHT19} introduced a similar trust-region-based triggering algorithm, as a general reinforcement learning, that has an extra rollback feature and possibly possesses some extra advantage. Nevertheless, the performance of SDGAT in transductive and inductive tasks also needs to be examined.

\appendices

\section {Tuning $\epsilon$ and $\delta$}
\begin{figure}[htb]
\centering
\includegraphics[width=7cm,height=6cm]{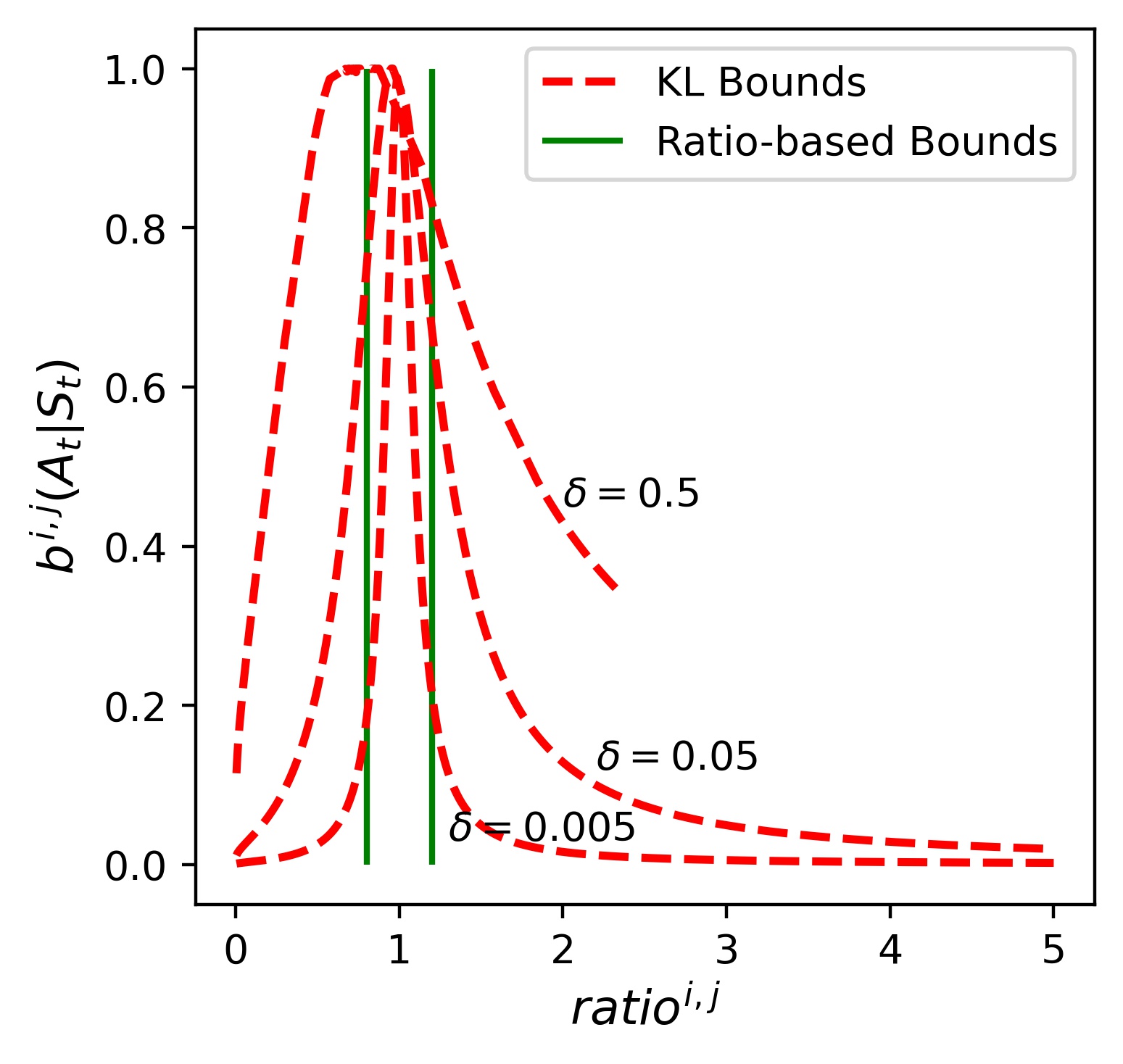}
\caption{Bounds with Different $\delta$}
\label{fig14} 
\end{figure}

TRC introduces a fundamental hyperparameter $\delta$, and the tuning of $(\epsilon-\delta)$ pairs becomes the concern. We tune the pairs based on our intuitive analysis of the KL bound gaps and grid search on possible pairs. With HYBRD and HYBRJ routines\cite{Powell1970}, we plot the KL bounds with $\delta$ increasing logarithmically and ratio-based bounds with $\epsilon=0.2$, as shown in Fig. \ref{fig14}, and we figure out that, neither 0.5 nor 0.005 is decent choice for $\delta$. When $\delta=0.5$, KL bounds nearly contain whole the ratio-based bounds, which at this point impose negligible constraint during clipping. On the other hand, when $\delta=0.005$, the constraint is over-strict, particularly when the action is favoured by the old policy, and therefore hampers the policy optimization process. Given analysis above, starting with 0.05, we grid-search on $\delta$ with increments or decrements of 0.015. 

Note that 0.1 and 0.2 are widely adopted choice for $\epsilon$, we focus our performance tests with their combinations with $\delta$. As an illustration, the grid search results on the CYTO dataset are shown in TABLE \ref{tab3}. They are evaluated by the mean and standard deviation of rewards during training along with the average SHD of the final results. The choice of $(\epsilon-\delta)$ pair for synthetic dataset is obtained likewise.

\begin{table}[tp]  
  
  \centering  
  \fontsize{9}{11}\selectfont  
  \caption{Grid-search Results on CYTO}  
    \begin{tabular}{|c|c|c|c|c|c|c|}  
    \hline  
    \multirow{2}{*}{Pairs}&  
    \multicolumn{3}{c|}{$\epsilon=0.1$}&\multicolumn{3}{c|}{$\epsilon=0.2$}\cr\cline{2-7}  
    &Mean&Std.&SHD&Mean&Std.&SHD\cr  
    \hline  
    \hline  
    $\delta=0.02$&-6.54&2.56&11.37&{\bf -6.19}&2.71&11.11\cr\hline  
    $\delta=0.035$&-6.25&{\bf 2.15}&{\bf 9.44}&-6.44&2.59&10.13\cr\hline  
    $\delta=0.05$&-6.69&3.27&10.88&-6.81&4.11&11.75\cr\hline  
    $\delta=0.065$&-6.55&3.46&12.44&-6.30&3.15&11.25\cr\hline  
    $\delta=0.08$&-6.84&3.71&13.11&-6.54&4.21&13.25\cr\hline  
    \end{tabular}
\label{tab3}  
\end{table}  

\section{Hyperparameters}
Table \ref{hyper} lists the hyperparameters adopted in this paper. They are given by their symbols and meanings, and are categorized into three groups: RL Procedural Parameters, SDGAT Configuration Parameters, RL Method Parameters.

\begin{table} [H]
\small
\caption{Hyperparameters}  
\begin{tabular}{m{1cm}<{\centering}m{4cm}<{\centering}m{3cm}<{\centering}}
\toprule[2pt]
\textbf{Symbol}&\textbf{Meaning}&\textbf{Value}\\
\hline
$\Lambda_1$&Upper Bound for $\lambda_1$& 0\\
\hline
$BIC_u$&Upper Bound for $\lambda_2$& -1\\
\hline
$\Delta_1$&Increasing Additive Increment for $\lambda_1$& 1\\
\hline
$\Delta_2$&Increasing Multiplicative Factor for $\lambda_2$& 10\\
\hline
$t_u$&Parameters Update Iterations& 1000\\
\toprule[2pt]
$n_s$&Number of Stacks in SDGAT& 6\\
\hline
$n_{gat}$&Number of First-hierarchical Heads& 4\\
\hline
$n_{sd}$&Number of Second-hierarchical Heads& 4\\
\hline
$d_k,d_v$&Hidden Dimensions for Encoder& 16,64\\
\hline
$d_h$&Hidden Dimension for Decoder& 16\\
\toprule[2pt]
$\alpha_{m}$&Moving Average Update Rate& 0.99\\
\hline
$\alpha_\theta,\alpha_\omega$&Learning Rates&0.001\\
\hline
$\lambda_e$&Entropy Term Weight&0.001\\
\hline
$S_b$&Experience Buffer Size&10*Batch Size\\
\end{tabular}
\label{hyper}
\end{table}

\section{Benefits of Moving Average}
\label{appc}

\begin{figure}[htb]
\centering
\includegraphics[width=9cm,height=7cm]{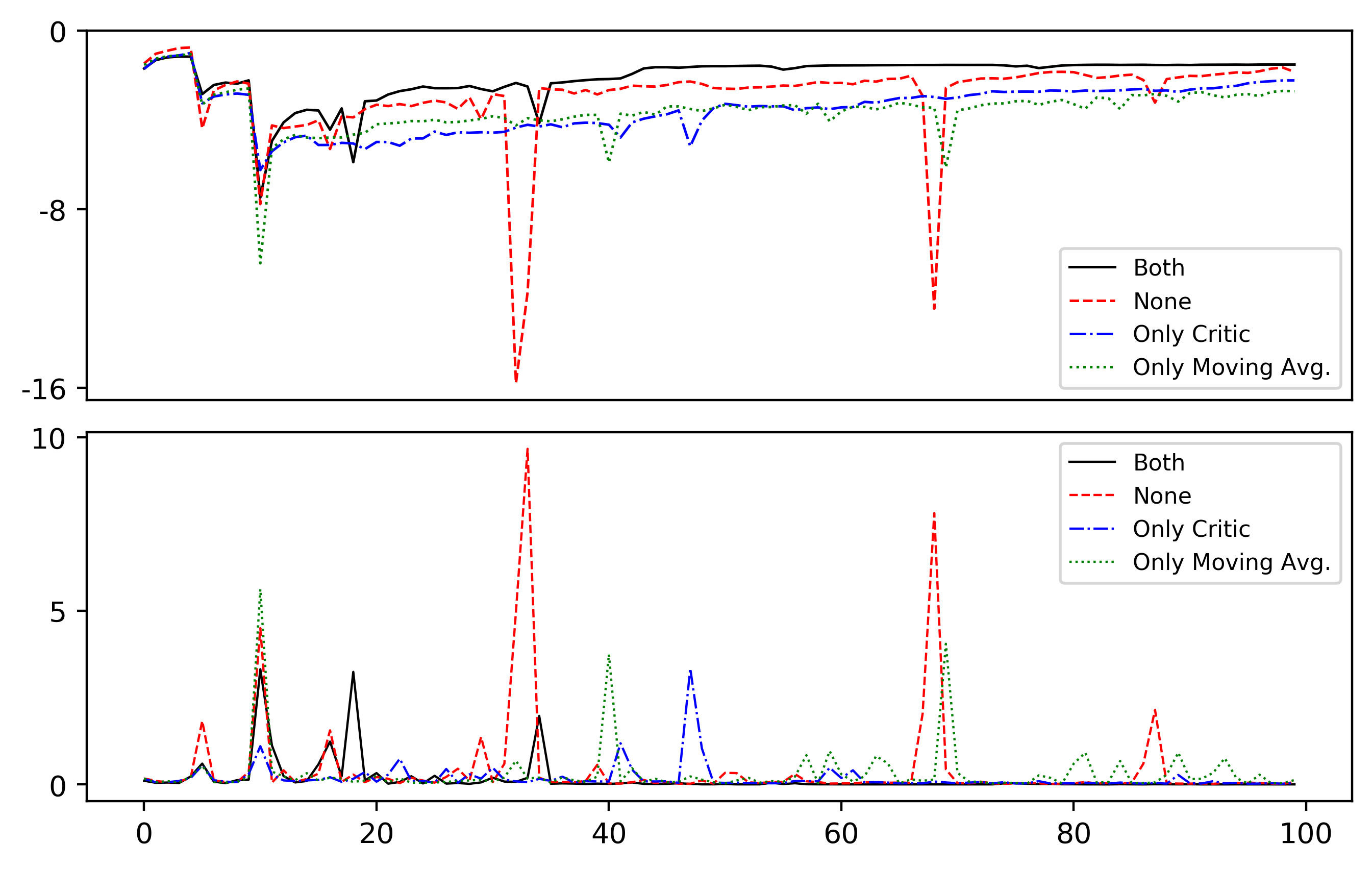}
\caption{Batch Rewards under Different Cases}
\label{fig15} 
\end{figure}

As there is no golden rule for baseline selection in current literature, we analyse the benefits of moving average with the experiment setting in \ref{exp1} under four cases: without baseline, with moving average baseline, with parametric baseline and with two baselines. The update direction of the gradient during optimization are equal among the four cases. We provide proof that, in the case with two baselines, the training objective is updated in the same gradient direction as in REINFORCE; the equality between other cases can be proved likewise:

\begin{equation}\nonumber
\begin{split}
&\because\sum_{a}{R_m\nabla\pi_\theta(a|S_t)}=R_m\nabla\sum_a{\pi_\theta(a|S_t)}=R_m\nabla1=0,\\
&\sum_{a}{V(S_t)\nabla\pi_\theta(a|S_t)}=V(S_t)\nabla\sum_a{\pi_\theta(a|S_t)}=V(S_t)\nabla1=0\\
&\therefore E[(R-R_m-V(S_t))\nabla\ln\pi_\theta(a|S_t)]\\
&=\sum_{a}{(R-R_m-V(S_t))\nabla\pi_\theta(a|S_t)}=\sum_{a}{R\nabla\pi_\theta(a|S_t)}\\
&=E[R\nabla\ln\pi_\theta(a|S_t)],
\end{split}
\end{equation}

We resort to the two-baseline method due to its empirical significance. The batch reward value is an indispensable optimization objective component and training indicator. Fig. \ref{fig15} shows the mean and standard deviation for batch rewards per 200 epochs under the four cases averaged over 20 training times. It can be inferred that by adopting moving average to support parametric baseline, optimization performance is least prone to suffer from degradation as the value of the batch reward remains stable and ends up with the highest mean reward.


\ifCLASSOPTIONcaptionsoff
  \newpage
\fi



\bibliographystyle{IEEEtran}
\bibliography{mybibfile}

\begin{thebibliography}{10}
\providecommand{\url}[1]{#1}
\csname url@samestyle\endcsname
\providecommand{\newblock}{\relax}
\providecommand{\bibinfo}[2]{#2}
\providecommand{\BIBentrySTDinterwordspacing}{\spaceskip=0pt\relax}
\providecommand{\BIBentryALTinterwordstretchfactor}{4}
\providecommand{\BIBentryALTinterwordspacing}{\spaceskip=\fontdimen2\font plus
\BIBentryALTinterwordstretchfactor\fontdimen3\font minus \fontdimen4\font\relax}
\providecommand{\BIBforeignlanguage}[2]{{%
\expandafter\ifx\csname l@#1\endcsname\relax
\typeout{** WARNING: IEEEtran.bst: No hyphenation pattern has been}%
\typeout{** loaded for the language `#1'. Using the pattern for}%
\typeout{** the default language instead.}%
\else
\language=\csname l@#1\endcsname
\fi
#2}}
\providecommand{\BIBdecl}{\relax}
\BIBdecl

\bibitem{Daniela2010}
D.~B. Fenker, M.~A. Schoenfeld, M.~R. Waldmann, H.~Sch{\"{u}}tze, H.~Heinze, and E.~D{\"{u}}zel, ``'virus and epidemic': Causal knowledge activates prediction error circuitry,'' \emph{J. Cogn. Neurosci.}, vol.~22, no.~10, pp. 2151--2163, 2010.

\bibitem{Bernhard2019}
B.~Sch{\"{o}}lkopf, ``Causality for machine learning,'' \emph{CoRR}, vol. abs/1911.10500, 2019.

\bibitem{Rainer2007}
R.~Opgen{-}Rhein and K.~Strimmer, ``From correlation to causation networks: a simple approximate learning algorithm and its application to high-dimensional plant gene expression data,'' \emph{{BMC} Syst. Biol.}, vol.~1, p.~37, 2007.

\bibitem{Peter2000}
P.~Spirtes, C.~Glymour, and R.~Scheines, \emph{Causation, Prediction, and Search, Second Edition}, ser. Adaptive computation and machine learning.\hskip 1em plus 0.5em minus 0.4em\relax {MIT} Press, 2000.

\bibitem{Chickering1995}
D.~M. Chickering, ``Learning bayesian networks is np-complete,'' in \emph{Learning from Data - Fifth International Workshop on {AISTATS} 1995. Proceedings}, D.~Fisher and H.~Lenz, Eds.\hskip 1em plus 0.5em minus 0.4em\relax Springer, 1995, pp. 121--130.

\bibitem{Zhu2020}
S.~Zhu, I.~Ng, and Z.~Chen, ``Causal discovery with reinforcement learning,'' in \emph{8th International Conference on Learning Representations, {ICLR} 2020}.\hskip 1em plus 0.5em minus 0.4em\relax OpenReview.net, 2020.

\bibitem{Schwarz1978}
G.~{Schwarz}, ``{Estimating the Dimension of a Model},'' \emph{Annals of Statistics}, vol.~6, no.~2, pp. 461--464, Jul. 1978.

\bibitem{Williams1992}
R.~J. Williams, ``Simple statistical gradient-following algorithms for connectionist reinforcement learning,'' \emph{Mach. Learn.}, vol.~8, pp. 229--256, 1992.

\bibitem{Sutton1999}
R.~S. Sutton, D.~A. McAllester, S.~P. Singh, and Y.~Mansour, ``Policy gradient methods for reinforcement learning with function approximation,'' in \emph{Advances in Neural Information Processing Systems 12}.\hskip 1em plus 0.5em minus 0.4em\relax The {MIT} Press, 1999, pp. 1057--1063.

\bibitem{Schulman2015}
J.~Schulman, S.~Levine, P.~Abbeel, M.~I. Jordan, and P.~Moritz, ``Trust region policy optimization,'' in \emph{Proceedings of the 32nd International Conference on Machine Learning, {ICML} 2015}, vol.~37.\hskip 1em plus 0.5em minus 0.4em\relax JMLR.org, 2015, pp. 1889--1897.

\bibitem{Schulman2017}
J.~Schulman, F.~Wolski, P.~Dhariwal, A.~Radford, and O.~Klimov, ``Proximal policy optimization algorithms,'' \emph{CoRR}, vol. abs/1707.06347, 2017.

\bibitem{Velickovic2017}
P.~Velickovic, G.~Cucurull, A.~Casanova, A.~Romero, P.~Li{\`{o}}, and Y.~Bengio, ``Graph attention networks,'' \emph{CoRR}, vol. abs/1710.10903, 2017.

\bibitem{Chow1968}
C.~K. Chow and C.~N. Liu, ``Approximating discrete probability distributions with dependence trees,'' \emph{{IEEE} Trans. Inf. Theory}, vol.~14, no.~3, pp. 462--467, 1968.

\bibitem{Tsamardinos2006}
I.~Tsamardinos, L.~E. Brown, and C.~F. Aliferis, ``The max-min hill-climbing bayesian network structure learning algorithm,'' \emph{Mach. Learn.}, vol.~65, no.~1, pp. 31--78, 2006.

\bibitem{Zheng2018}
X.~Zheng, B.~Aragam, P.~Ravikumar, and E.~P. Xing, ``Dags with {NO} {TEARS:} continuous optimization for structure learning,'' in \emph{NeurIPS 2018}, 2018, pp. 9492--9503.

\bibitem{Huang2018}
B.~Huang, K.~Zhang, and Y.~Lin, ``Generalized score functions for causal discovery,'' in \emph{Proceedings of the 24th {ACM} {SIGKDD}}.\hskip 1em plus 0.5em minus 0.4em\relax {ACM}, 2018, pp. 1551--1560.

\bibitem{Peters2014}
J.~Peters, J.~M. Mooij, D.~Janzing, and B.~Sch{\"{o}}lkopf, ``Causal discovery with continuous additive noise models,'' \emph{J. Mach. Learn. Res.}, vol.~15, no.~1, pp. 2009--2053, 2014.

\bibitem{Ahmed2019AML}
M.~U. Ahmed, S.~Brickman, A.~Dengg, N.~Fasth, M.~Mihajlovi{\'c}, and J.~D. Norman, ``A machine learning approach to classify pedestrians’ event based on imu and gps,'' 2019.

\bibitem{9262059}
A.~{Belhadi}, Y.~{Djenouri}, D.~{Djenouri}, T.~{Michalak}, and J.~C.~W. {Lin}, ``Deep learning versus traditional solutions for group trajectory outliers,'' \emph{IEEE Transactions on Cybernetics}, pp. 1--12, 2020.

\bibitem{preitl2007}
S.~Preitl, R.-E. Precup, P.~Zsuzsa, S.~Vaivoda, S.~Kilyeni, and J.~Tar, ``Iterative feedback and learning control. servo systems applications,'' \emph{IFAC Proceedings Volumes (IFAC-PapersOnline)}, vol.~1, pp. 16--27, 01 2007.

\bibitem{KatriniokA16}
A.~Katriniok and D.~Abel, ``Adaptive ekf-based vehicle state estimation with online assessment of local observability,'' \emph{{IEEE} Trans. Control. Syst. Technol.}, vol.~24, no.~4, pp. 1368--1381, 2016.

\bibitem{Goudet2017}
O.~{Goudet}, D.~{Kalainathan}, P.~{Caillou}, I.~{Guyon}, D.~{Lopez-Paz}, and M.~{Sebag}, ``Learning functional causal models with generative neural networks,'' Sep. 2017.

\bibitem{Kalainathan2018}
D.~{Kalainathan}, O.~{Goudet}, I.~{Guyon}, D.~{Lopez-Paz}, and M.~{Sebag}, ``{Structural Agnostic Modeling: Adversarial Learning of Causal Graphs},'' \emph{arXiv e-prints}, Mar. 2018.

\bibitem{Yu2019}
Y.~Yu, J.~Chen, T.~Gao, and M.~Yu, ``{DAG-GNN:} {DAG} structure learning with graph neural networks,'' in \emph{{ICML} 2019}, 2019.

\bibitem{Gori2005}
M.~{Gori}, G.~{Monfardini}, and F.~{Scarselli}, ``A new model for learning in graph domains,'' in \emph{IEEE International Joint Conference on Neural Networks, 2005}, vol.~2, 2005, pp. 729--734 vol. 2.

\bibitem{Scarselli2009}
F.~Scarselli, M.~Gori, A.~C. Tsoi, M.~Hagenbuchner, and G.~Monfardini, ``The graph neural network model,'' \emph{{IEEE} Trans. Neural Networks}, vol.~20, no.~1, pp. 61--80, 2009.

\bibitem{Tang2019}
S.~Tang, B.~Li, and H.~Yu, ``Chebnet: Efficient and stable constructions of deep neural networks with rectified power units using chebyshev approximations,'' \emph{CoRR}, vol. abs/1911.05467, 2019.

\bibitem{Hamilton2017}
W.~L. Hamilton, R.~Ying, and J.~Leskovec, ``Inductive representation learning on large graphs,'' \emph{CoRR}, vol. abs/1706.02216, 2017.

\bibitem{9043893}
T.~T. {Nguyen}, N.~D. {Nguyen}, and S.~{Nahavandi}, ``Deep reinforcement learning for multiagent systems: A review of challenges, solutions, and applications,'' \emph{IEEE Transactions on Cybernetics}, vol.~50, no.~9, pp. 3826--3839, 2020.

\bibitem{Polydoros2017}
A.~S. Polydoros and L.~Nalpantidis, ``Survey of model-based reinforcement learning: Applications on robotics,'' \emph{J. Intell. Robotic Syst.}, vol.~86, no.~2, pp. 153--173, 2017.

\bibitem{8613842}
Y.~{Wen}, J.~{Si}, A.~{Brandt}, X.~{Gao}, and H.~H. {Huang}, ``Online reinforcement learning control for the personalization of a robotic knee prosthesis,'' \emph{IEEE Transactions on Cybernetics}, vol.~50, no.~6, pp. 2346--2356, 2020.

\bibitem{Wang2019}
Y.~Wang, L.~Zhang, L.~Wang, and Z.~Wang, ``Multitask learning for object localization with deep reinforcement learning,'' \emph{{IEEE} Trans. Cogn. Dev. Syst.}, vol.~11, no.~4, pp. 573--580, 2019.

\bibitem{8906005}
J.~{Li}, J.~{Ding}, T.~{Chai}, and F.~L. {Lewis}, ``Nonzero-sum game reinforcement learning for performance optimization in large-scale industrial processes,'' \emph{IEEE Transactions on Cybernetics}, vol.~50, no.~9, pp. 4132--4145, 2020.

\bibitem{Zoph2016}
B.~{Zoph} and Q.~V. {Le}, ``{Neural Architecture Search with Reinforcement Learning},'' \emph{arXiv e-prints,}, Nov. 2016.

\bibitem{Bollen1989}
K.~A. Bollen, \emph{Structural Equations with Latent Variables}.\hskip 1em plus 0.5em minus 0.4em\relax Wiley, 1989.

\bibitem{Spirtes2000}
P.~Spirtes, C.~Glymour, and R.~Scheines, \emph{Causation, Prediction, and Search, Second Edition}, ser. Adaptive computation and machine learning.\hskip 1em plus 0.5em minus 0.4em\relax {MIT} Press, 2000.

\bibitem{Shimizu2006}
S.~Shimizu, P.~O. Hoyer, A.~Hyv{\"{a}}rinen, and A.~J. Kerminen, ``A linear non-gaussian acyclic model for causal discovery,'' \emph{J. Mach. Learn. Res.}, vol.~7, pp. 2003--2030, 2006.

\bibitem{Cho2014}
K.~{Cho}, B.~{van Merrienboer}, D.~{Bahdanau}, and Y.~{Bengio}, ``{On the Properties of Neural Machine Translation: Encoder-Decoder Approaches},'' \emph{arXiv e-prints,}, Sep. 2014.

\bibitem{Nallapati2016}
R.~{Nallapati}, B.~{Zhou}, C.~{Nogueira dos santos}, C.~{Gulcehre}, and B.~{Xiang}, ``{Abstractive Text Summarization Using Sequence-to-Sequence RNNs and Beyond},'' \emph{arXiv e-prints,}, Feb. 2016.

\bibitem{Williams1991}
R.~J. Williams and J.~Peng, ``Function optimization using connectionist reinforcement learning algorithms,'' \emph{Connection Science}, vol.~3, pp. 241--268, 1991.

\bibitem{Schaul2015}
T.~{Schaul}, J.~{Quan}, I.~{Antonoglou}, and D.~{Silver}, ``{Prioritized Experience Replay},'' \emph{arXiv e-prints,}, Nov. 2015.

\bibitem{Ilyas2018}
A.~Ilyas, L.~Engstrom, S.~Santurkar, D.~Tsipras, F.~Janoos, L.~Rudolph, and A.~Madry, ``Are deep policy gradient algorithms truly policy gradient algorithms?'' \emph{CoRR}, vol. abs/1811.02553, 2018.

\bibitem{Ramsey2017}
J.~Ramsey, M.~Glymour, R.~Sanchez{-}Romero, and C.~Glymour, ``A million variables and more: the fast greedy equivalence search algorithm for learning high-dimensional graphical causal models, with an application to functional magnetic resonance images,'' \emph{Int. J. Data Sci. Anal.}, vol.~3, no.~2, pp. 121--129, 2017.

\bibitem{Buhlmann2013}
P.~B{\"{u}}hlmann, J.~Peters, and J.~Ernest, ``{CAM:} causal additive models, high-dimensional order search and penalized regression,'' \emph{CoRR}, vol. abs/1310.1533, 2013.

\bibitem{Lachapelle2019}
S.~Lachapelle, P.~Brouillard, T.~Deleu, and S.~Lacoste{-}Julien, ``Gradient-based neural {DAG} learning,'' \emph{CoRR}, vol. abs/1906.02226, 2019.

\bibitem{Sachs2005}
K.~{Sachs}, O.~{Perez}, D.~{Pe'er}, D.~A. {Lauffenburger}, and G.~P. {Nolan}, ``{Causal Protein-Signaling Networks Derived from Multiparameter Single-Cell Data},'' \emph{Science}, vol. 308, no. 5721, pp. 523--529, Apr. 2005.

\bibitem{BulckeLNRMVMM06}
T.~V. den Bulcke, K.~V. Leemput, B.~Naudts, P.~van Remortel, H.~Ma, A.~Verschoren, B.~D. Moor, and K.~Marchal, ``Syntren: a generator of synthetic gene expression data for design and analysis of structure learning algorithms,'' \emph{{BMC} Bioinform.}, vol.~7, p.~43, 2006.

\bibitem{WangHT19}
Y.~Wang, H.~He, and X.~Tan, ``Truly proximal policy optimization,'' in \emph{{UAI} 2019}, ser. Proceedings of Machine Learning Research, vol. 115.\hskip 1em plus 0.5em minus 0.4em\relax {AUAI} Press, pp. 113--122.

\bibitem{Powell1970}
M.~J.~D. Powell, ``A hybrid method for nonlinear equations,'' in \emph{Numerical Methods for Nonlinear Algebraic Equations}, P.~Rabinowitz, Ed.\hskip 1em plus 0.5em minus 0.4em\relax Gordon and Breach, 1970.

\end{thebibliography}
%
%
%

%

\begin{IEEEbiography}[{\includegraphics[width=1in,height=1.25in,clip,keepaspectratio]{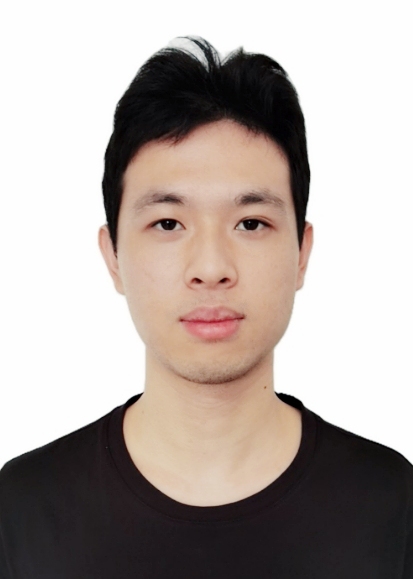}}]
{Shixuan Liu}
received the B.S. degree in Systems Engineering in 2019 from National University of Defense Technology, Changsha, P.R. China. He is currently working toward the Ph.D degree with National University of Defense Technology. His research interests include reinforcement learning, causal discovery and computer vision.
\end{IEEEbiography}

\vspace{-20pt}
\begin{IEEEbiography}[{\includegraphics[width=1in,height=1.25in,clip,keepaspectratio]{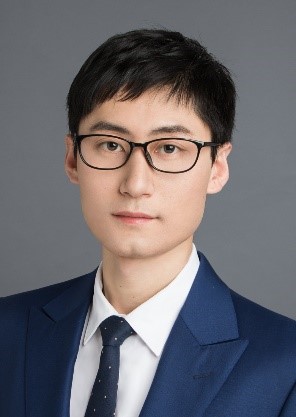}}]
{Yanghe Feng}
received the master's and PhD degrees from Information System and Engineering Laboratory in the National University of Defense Technology. He is a associate professor in the National University of Defense Technology. His primary research interests include the casual discovery and inference, active learning and reinforcement learning. His PhD research focused on building ''the plan online and learn offline'' framework to enable computers with the abilities to analyze, recognize and predict real-world uncertainty.
\end{IEEEbiography}

\vspace{-20pt}
\begin{IEEEbiography}[{\includegraphics[width=1in,height=1.25in,clip,keepaspectratio]{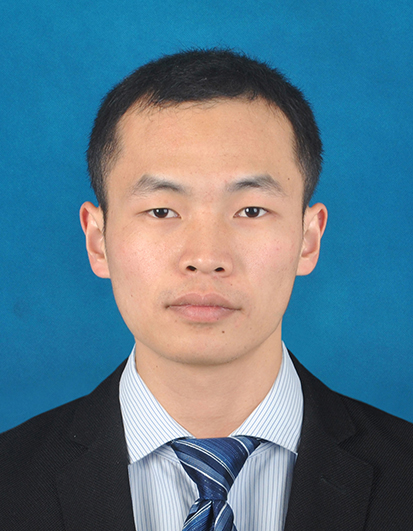}}]
{Keyu Wu}
received the Ph.D. degree in electrical engineering from the University of Alberta, Edmonton, AB, Canada, in 2018. He is currently an assistant professor in National University of Defense Technology. His primary research interests include dynamic stochastic control, reinforcement learning, and their applications in networked systems.
\end{IEEEbiography}

\vspace{-20pt}
\begin{IEEEbiography}[{\includegraphics[width=1in,height=1.25in,clip,keepaspectratio]{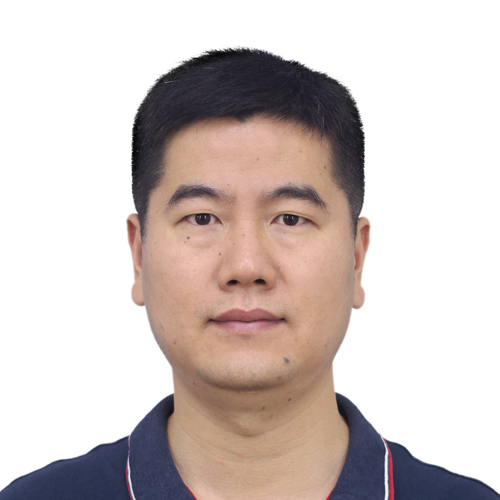}}]
{Guangquan Cheng}
received the M.Sc.and Ph.D. degrees from the National University of Defense Technology, Changsha, China, in 2005 and 2010, respectively, where he is currently an Associate Professor. His current research interests include complex network analysis and decision-making support technology.
\end{IEEEbiography}

\vspace{-20pt}
\begin{IEEEbiography}[{\includegraphics[width=1in,height=1.25in,clip,keepaspectratio]{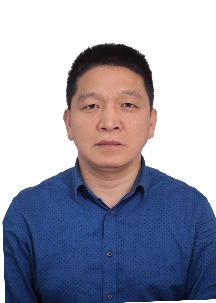}}]
{Jincai Huang}
is a Professor of the National University of Defense Technology, Changsha, Hunan, China, and a researcher of Science and Technology on Information Systems Engineering Laboratory. His main research interests include artificial general intelligence, deep reinforcement learning, and multi-agent systems.
\end{IEEEbiography}

\vspace{-20pt}
\begin{IEEEbiography}[{\includegraphics[width=1in,height=1.25in,clip,keepaspectratio]{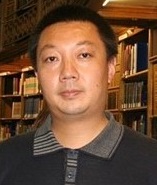}}]
{Zhong Liu}
received the Ph.D. degree in management science from National University of Defense Technology (NUDT), Changsha, China, in 2000. He is currently a Professor with NUDT. He is also vice-dean with the College of Systems Engineering Laboratory, NUDT, and a Senior Advisor with the Research Center for Computational Experiments and Parallel Systems, NUDT. His main research interests include planning systems, computational organization, and intelligent systems.
\end{IEEEbiography}
\vfill



\end{document}